\pdfoutput=1

\documentclass[11pt, usenames,dvipsnames]{article}

\usepackage[]{emnlp2021}

\usepackage{times}
\usepackage{latexsym}
\usepackage{verbatim}
\usepackage[textsize=scriptsize]{todonotes}
\usepackage{hyperref}
\usepackage{bm}
\usepackage{booktabs}
\usepackage{multirow}
\usepackage{siunitx}
\usepackage{amsmath, amssymb}
\usepackage{dsfont}
\usepackage{xcolor}
\definecolor{Red0}{RGB}{255, 245, 245}
\definecolor{Red1}{RGB}{255, 230, 230}
\definecolor{Red2}{RGB}{255, 215, 215}
\definecolor{Red3}{RGB}{255, 200, 200}
\definecolor{Red4}{RGB}{255, 185, 185}
\definecolor{Red5}{RGB}{255, 170, 170}

\definecolor{Green0}{RGB}{245, 255, 245}
\definecolor{Green1}{RGB}{230, 255, 230}
\definecolor{Green2}{RGB}{215, 255, 215}
\definecolor{Green3}{RGB}{200, 255, 200}
\definecolor{Green4}{RGB}{185, 255, 185}

\usepackage[T1]{fontenc}

\usepackage[utf8]{inputenc}

\usepackage{microtype}

\newcommand{\liyan}[1]{\textcolor{black}{#1}}

\title{Making Document-Level Information Extraction\\Right for the Right Reasons}

\author{Liyan Tang$^{1}$,  Dhruv Rajan$^{1}$, Suyash Mohan$^{2}$, Abhijeet Pradhan$^{3}$, R. Nick Bryan$^{1}$,  Greg Durrett$^{1}$ \\
        $^{1}$The University of Texas at Austin \\ $^{2}$University of Pennsylvania\\ $^{3}$Galileo CDS Inc. \\
        \texttt{lytang@utexas.edu, dhruv.rajan@utexas.edu } \\\texttt{Suyash.Mohan@pennmedicine.upenn.edu, ap@galileocds.com} \\ \texttt{nick.bryan@austin.utexas.edu, gdurrett@cs.utexas.edu}
        }

\begin{document}
\maketitle
\begin{abstract}
Document-level models for information extraction tasks like slot-filling are flexible: they can be applied to settings where information is not necessarily localized in a single sentence. For example, key features of a diagnosis in a radiology report may not be explicitly stated in one place, but nevertheless can be inferred from parts of the report's text. However, these models can easily learn spurious correlations between labels and irrelevant information. This work studies how to ensure that these models make correct inferences from complex text \textbf{and} make those inferences in an auditable way: beyond just being right, are these models ``right for the right reasons?'' We experiment with post-hoc evidence extraction in a predict-select-verify framework using feature attribution techniques. We show that regularization with small amounts of evidence supervision during training can substantially improve the quality of extracted evidence. We evaluate on two domains: a small-scale labeled dataset of brain MRI reports and a large-scale modified version of DocRED \cite{Yao2019} and show that models' plausibility can be improved with no loss in accuracy.\footnote{\liyan{Code available at \url{https://github.com/Liyan06/DocumentIE}.}}
\end{abstract}

\section{Introduction}

Document-level information extraction \citep{Yao2019, Christopoulou2019, Xiao2020, nan2020lsr} has seen great strides due to the rise of pre-trained models \cite{Devlin2019}. But in high-stakes domains like medical information extraction \citep{Irvin2019, Matthew2020, Smit2020}, machine learning models are still too error-prone to use broadly. Since they are not perfect, they typically play the role of assisting users in tasks like building cohorts \cite{Pons_2016_nlp_in_rad} or in providing clinical decision support \cite{DemnerFushman_cds}.

\liyan{To be most usable in conjunction with users, these systems should not just produce a decision, but a justification for their answer. The ideal system therefore obtains high predictive accuracy, but also returns a rationale that allows a human to verify the predicted label \cite{RudieEtAl2019}.}

\begin{figure}
    \centering
    \includegraphics[scale=0.75,trim=0 89mm 0mm 27mm]{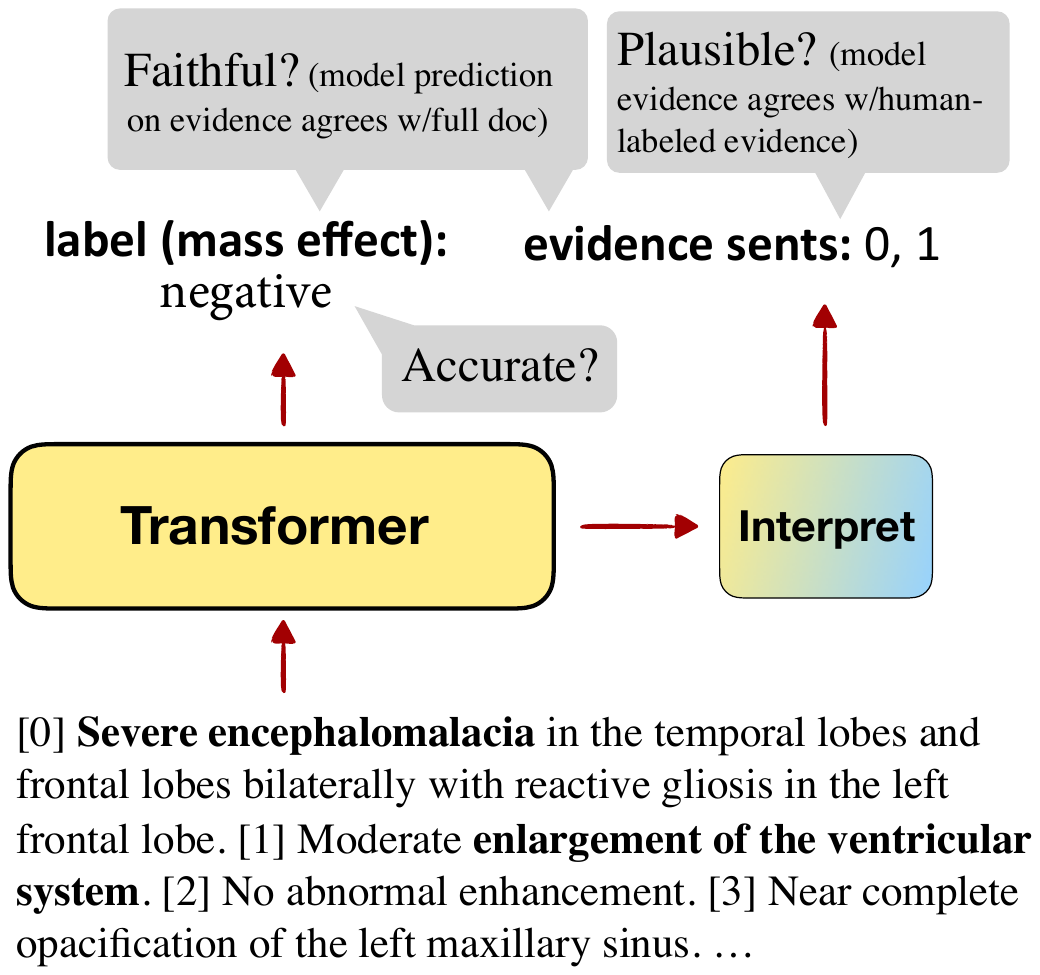}
    \caption{Our basic model setup. A Transformer-based model makes document-level predictions on an example of our brain MRI reports. An interpretation method extracts the evidence sentences used by the model. Our system is evaluated according to the criteria of accuracy, faithfulness, and plausibility.
    }
    \label{fig:architecture}
\end{figure}

Our goal is to study document-level information extraction systems that are both accurate and which make predictions based on the correct information \citep{DoshiVelez2017}. This process involves identifying what evidence the model actually used, verifying the model's prediction based on that evidence, and checking whether that evidence aligns with what humans would use, which would allow a user to quickly see if the system is correct. For example, in Figure~\ref{fig:architecture}, localizing the prediction of \emph{mass effect} (a feature expressing whether there is evidence of brain displacement by a mass like a tumor) to the first two sentences allows a trained user in a clinical decision support setting to easily verify what was extracted here. Our evidence extraction hews to principles of both faithfulness and plausibility \citep{Jain2020, Jacovi2020, Miller2019}. 

Rather than use complex approaches with intermediate latent variables for extraction \cite{Lei2016}, we focus on what can be done with off-the-shelf pre-trained models \cite{liu2019} using post-hoc interpretation. We explore various interpretation methods to find key parts of each document that were used by the model. We ask two questions: first, can we identify the document sentences that truly contributed to the prediction (faithfulness)? Using the ranking of sentences provided by an interpretation method, we extract a set of sentences where the model returns nearly the same prediction as before, thus \emph{verifying} that these sentences are a sufficient explanation for the model. Second, do these document sentences align with what users annotated (plausibility)? Unsurprisingly, we find that this alignment is low in a basic Transformer model.


To further improve the alignment with human annotation, we consider injecting small amounts of sentence-level supervision. Critically, in the brain MRI extraction setting we consider (see Table \ref{ReportExample}), large-scale sentence-level annotation is not available; most instances in the dataset only have document-level labels from existing clinical decision support systems, making it a weakly-supervised setting \citep{Pruthi2020-2, Patel2020}. We explore two methods for using this small amount of annotation, chiefly based around supervising or regularizing the model's behavior. One notion is entropy maximization: the model should be uncertain when it isn't exposed to sufficient evidence \citep{Feng2019}. Another is attention regularization where the model is encouraged to attend to key pieces of evidence. While attention is not entirely connected with what the model uses \citep{Jain2019}, we can investigate whether this leads to a model whose explanations leverage this information more heavily. 

We validate our methods first on a small dataset of radiologists' observations from brain MRIs. These reports are annotated with document-level key features related to different aspects of the report, which we want to extract in a faithful way. We see positive results here even in a small-data condition, but to understand how this method would scale with larger amounts of data, we adapt the DocRED relation extraction task \cite{Yao2019} to be a document-level classification task. The question of which sentence in the document describes the relation between the two entities, if there even is one, is still quite challenging, and we show our techniques can lead to improvements in a weakly-labeled setting here as well.

Our contributions are (1) We apply evidence extraction methods to document-level classification and slot-filling tasks, emphasizing a new brain MRI dataset that we annotate. (2) We explore using weak sentence-level supervision in two techniques adapted from prior work; (3) We evaluate pre-trained models and evidence extraction through various interpretation methods for plausibility compared to human annotation, while ensuring faithfulness of the evidence.


    
    
    
    

\begin{table}[]
    \small
    \begin{tabular}{p{0.94\linewidth}} \toprule
    \multicolumn{1}{c}{\textbf{Report Finding}} \\
         \emph{[0]} \textbf{\textcolor{Red}{Severe encephalomalacia}} in the temporal lobes and frontal lobes \textbf{\textcolor{Orange}{bilaterally}} with reactive \textbf{\textcolor{RoyalBlue}{gliosis}} in the left frontal lobe. \emph{[1]} Moderate \textbf{\textcolor{Red}{enlargement of the ventricular system}}. \emph{[2]} \textbf{\textcolor{OliveGreen}{No abnormal enhancement}}. \emph{[3]} Near complete opacification of the left maxillary sinus. ...\\
         \\
         \begin{tabular}{llr}
            \textcolor{Red}{\textbf{mass\_effect}}: negative &\textbf{evid}: \emph{[0, 1]} & implicit\\
            \textcolor{Orange}{\textbf{side}}: bilateral &\textbf{evid}: \emph{[0]}	& explicit\\
            \textcolor{RoyalBlue}{\textbf{t2}}: increased &\textbf{evid}: \emph{[0]} & implicit\\
            \textcolor{OliveGreen}{\textbf{contrast\_enhancement}}: No &\textbf{evid}: \emph{[2]} & explicit \\
        \end{tabular} \\
    \bottomrule
    \end{tabular}
    \caption{Example from annotated brain MRI reports. Labels and supporting evidence for $4$ key features are annotated for this example report presented. ``Explicit'' means the label of given key feature can be directly inferred by the highlighted terms; ``implicit'' instead indicates that it requires domain knowledge and potential reasoning skills to label. We want the model to identify implicit features while not leveraging dataset biases or reasoning incorrectly about explicit ones.}
    \label{ReportExample}
\end{table}

\section{Background} \label{sec:background}

\subsection{Motivation}

We start with an example from a brain MRI report in Table~\ref{ReportExample}. Medical information extraction involves tasks such as identifying important medical terms from text \citep{Irvin2019, Smit2020} and normalizing names into standard concepts using domain-specific ontologies \citep{Cho2017}. One application in clinical decision support, shown here, requires extracting the values of certain key features (clinical findings) from these reports or medical images \cite{Rudie2021, Duong2019}. This extraction should be accurate, but it should also make predictions that are correctly sourced, to facilitate review by a radiologist or someone else using the system \cite{Rauschecker2020, Cook2018}.

The finding section of a brain MRI report often describes these key features in both explicit and implicit ways. For instance, contrast enhancement, one of our key features, is mentioned explicitly much of the time; see \emph{no abnormal enhancement} in the third sentence. A rule-based system can detect this type of evidence easily. But some key features are harder to identify and require reasoning over context and draw on implicit cues. For example, \emph{severe encephalomalacia} in the first sentence and \emph{enlargement of the ventricular system} in the following sentence are both implicit signs of positive mass effect and either is sufficient to infer the label. It is significantly harder to built a rule-based extractor for this case. Learning-based systems have the potential to do much better here, but lack of understanding about their behavior can lead to hard-to-predict failure modes, such as acausal prediction of key features (e.g., inferring evidence about mass effect from a hypothesized diagnosis somewhere in the report, where the causality is backwards).

Our work aims to leverage the ability of learning-based systems to capture implicit features while improving their ability to make predictions that are sourced from the correct evidence and can be easily verified. 

\subsection{Problem Setting}

The problem we tackle in this work can be viewed as document-level classification.  Let $D = \{x_1, \ldots, x_n\}$ be a document consisting of $n$ sentences. The document is annotated with a set of labels $(t_i,y_i)$ where $t_i$ is an auxiliary input specifying a particular task for this document (e.g., mass effect) and $y_i$ is the label associated with that task from a discrete label space $\{1,\ldots,d\}$. In our adaptation of the DocRED task, we consider  $t = (e_1,e_2)$ to classify the relationship (if any) between a pair of entities $(e_1,e_2)$ in a document, defined in Section \ref{subsection: Docred}.

Our method takes a pair $(D,t)$ and then computes the label $\hat{y}_t$ from a predictor $\hat{y}_t = f(D,t)$. We can then extract \textbf{evidence}, a set of sentences, post-hoc using a separate procedure $g$ such as a feature attribution method: $\hat{E}_t = g(f,D,t)$

\paragraph{Supervision} In addition to the labels $y_t$, we assume access to a small number of examples with additional supervision in each domain. That is, for a $(D,t,y_t)$ triple, we also assume we are given a set $E = \{x_{i_1},\ldots, x_{i_m}\}$ of ground-truth evidence with sentence indices $\{i_1,\ldots,i_m\}$. This evidence should be sufficient to compute the label, but not always necessary; for example, if multiple sentences can contribute to the prediction, they might all be listed as supporting evidence here. See Section~\ref{sec:improving_ee} for more details.

\subsection{Related Work}

Our work fits into a broader thread of work on information extraction with partial annotation \citep{Han2020}. Due to the cost of collecting large-scale data with good quality, distant supervision (DS) \citep{mintz2009} and ways to denoise auto-labeled data from DS  \citep{surdeanu2012, wang2018} have been widely explored. However, the sentence-level setting typically features much less ambiguity about evidence needed to predict a relation compared to the document-level setting we explore. Several document-level RE datasets \citep{Li2016-2, Peng2017} have been proposed as well as efforts to tackle these tasks \citep{Christopoulou2019, Xiao2020, nan2020lsr}, which we explicitly build from.

\paragraph{Explanation techniques} To identify the sentences that the model considers as evidence, we draw on a recent body of work in explainable NLP focused on identifying salient features of the input. These primarily consist of input attribution techniques, such as LIME \citep{Ribeiro2016}, input reductions \citep{Li2016-1, Feng2018}, attention-based explanations \citep{Bahdanau2015} and gradient-based methods \citep{Simonyan2014, Selvaraju2017, sundararajan2017, Shrikumar2017}. \liyan{In present work, we extract rationales using commonly used model interpretation methods (described in Section~\ref{sec:evi_extract}) and focus on doing a thorough evaluation of the capabilities of DeepLIFT \citep{Shrikumar2017} given its competitive performance in our interpretation methods comparison (Appendix~\ref{sec:interpret_compare}).}


\paragraph{Frameworks for interpretable pipelines} Our goal of building a system grounded in evidence draws heavily on recent work on attribution techniques and model explanations, particularly notions of faithfulness and plausibility. \emph{Faithfulness} refers to how accurately the explanation provided by the model truly reflects the information it used in the reasoning process \citep{Jain2020}. On the other hand, \emph{plausibility} indicates to what extent the interpretation provided by the model makes sense to a person.\footnote{The ERASER benchmark \cite{DeYoung2020} is a notable recent effort to evaluate explanation plausibility. However, we do not consider it here; we focus on the document-level classification setting, and many of the ERASER tasks are not suitable or relevant for the approaches we consider, either being not natural (FEVER) or not having the same challenges as document-level classification.}

``\emph{Select-then-predict}'' approaches are one way to enforce faithfulness in pipelines \cite{Jain2020}: important snippets from inputs are extracted and passed through a classifier to make predictions. Past work has used hard \citep{Lei2016} or soft \citep{Zhang2016} rationales, and other work has explicitly looked at tradeoffs in the amount of text extracted \citep{Paranjape2020}. 

\citet{Jacovi2020} note several problems with this setup. Our work aims to align model behavior with what cues we expect a model to use (plausibility), but uses the predict-select-verify paradigm \cite{Jacovi2020} to ensure that these are actually sufficient cues for the model. Like our work, \citet{Pruthi2020-2} simultaneously trained a BERT-based model \citep{Devlin2019} for the prediction task and a linear-CRF \citep{Lafferty2001} module on top of it for the evidence extraction task with shared parameters. Compared to their work, we focus explicitly on what can be done with pre-trained models alone, not augmenting the model for evidence extraction. 


\section{Methods}

The systems we devise take $(D,t)$ pairs as input and return (a) predicted labels $\hat{y}_t$ for each $t$; (b) sets of extracted evidence sentences $\hat{E}_t$ from an interpretation method. Figure~\ref{fig:architecture} shows the basic setting.

\subsection{Transformer Classification Model}

We use RoBERTa \citep{liu2019} as our document classifier. \liyan{RoBERTa is a strong method that holds up even against more recent baselines with architectures designed for DocRED \cite{zhou2021}.}
For each of our two domains, we use different pre-trained weights, as described in the training details in Appendix \ref{section:trainingdetails}. The task inputs are described in Section~\ref{sec:data_eval}.

\subsection{Interpretation for Evidence Extraction} \label{sec:evi_extract}

\liyan{Given any interpretation method as well as our model $\hat{y}_t = f(D,t)$}, we compute attribution scores with respect to the predicted class $y_t$ for each token in the RoBERTa input representation. We then average over the absolute value of attribution score for each token in that sentence to give sentence-level scores $\{s_1,\ldots,s_n\}$. These give us a ranking of the sentences. Given a fixed number of evidence sentences $k$ to extract, we can extract the top $k$ sentences by these scores.

\liyan{We experiment with the following four widely used interpretation techniques in the present work. \textbf{LIME} \cite{Ribeiro2016} offers explanations of an input by approximating the model’s predictions locally with an interpretable model. \textbf{Input Gradient} \cite{hechtlinger2016} and \textbf{Integrated Gradients} \cite{sundararajan2017} use gradients of the label with respect to the input to assess input importance; Integrated Gradients approximates the integral of this gradient with respect to the input along a straight path from a reference baseline.\footnote{We use the most typical baseline that consists of replacing the inputs in $D$ with {\fontfamily{qcr}\selectfont [MASK]} tokens from RoBERTa.} \textbf{DeepLIFT} \citep{Shrikumar2017} attributes the change in the output from a reference output in terms of the difference in input from the reference input.   Unless stated otherwise, we use DeepLIFT as our interpretation method, since it achieves the best results (comparable to Input Gradient) among the four interpretation options. Full comparison of interpretation methods is in Appendix~\ref{sec:interpret_compare}.}


To verify the extracted evidence \cite{Jacovi2020}, our main technique (\textsc{Sufficient}) feeds the model increasingly larger subsets of the document ranked by attribution scores (e.g., first $\{s_\textrm{max}\}$, then $\{s_\textrm{max}, s_\textrm{2nd-max}\}$, etc.) until it (a) makes the same prediction as when taking the whole document as input and (b) assigns that prediction at least $\lambda$ times the probability\footnote{The value of $\lambda$ is a tolerance hyper-parameter for selecting sentences and it set to $0.8$ throughout the experiments. \liyan{Our method is robust to the choice of $\lambda$ in a reasonable range, as shown in Appendix~\ref{sec:interpret_compare}}.} when the whole document is taken as input. We consider this attribution faithful: it is a subset of the input supporting the model's decision judged as important by the attribution method. 

\subsection{Improving Evidence Extraction}
\label{sec:improving_ee}

While many document-level extraction settings do not have sentence-level attributions labeled for every decision, one can in practice annotate a small fraction of a dataset with such ground-truth rationales. This is indeed the case for our brain MRI case study. Past work has shown significant benefits from integrating this supervision into learning \cite{strout-etal-2019-human,DuaEtAl2020,Pruthi2020-1}.

\begin{figure}
    \small
    \centering
    \includegraphics[scale=0.19,trim=0 30mm 0mm 30mm]{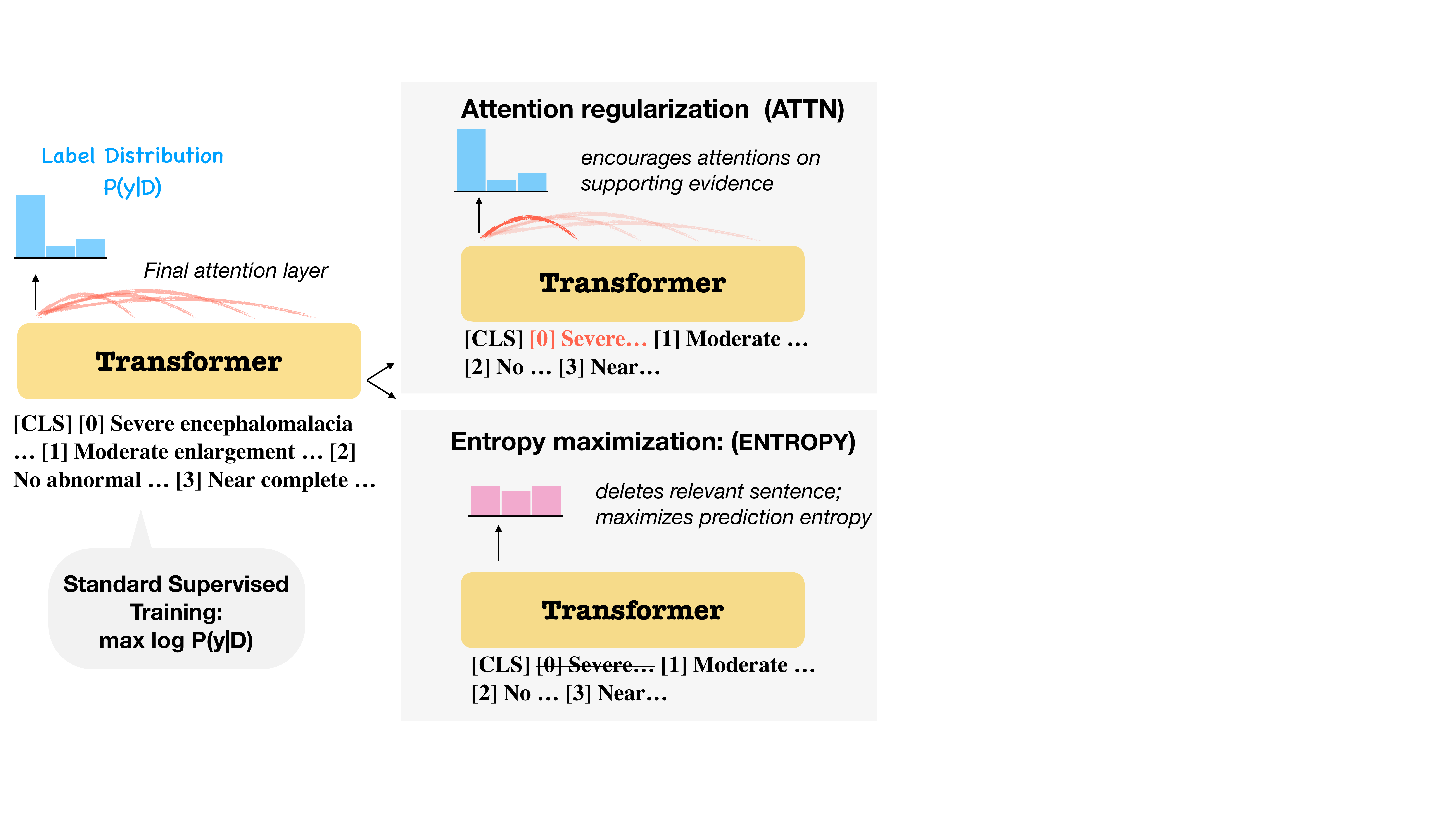}
    \caption{An illustration of attention regularization and entropy maximization using the example in Table \ref{ReportExample}. The model is predicting the label for key feature \emph{\text{t2}}.}
    \label{fig:illustraction}
\end{figure}

Assume that a subset of our labeled data consists of $(D, t, y_t, E_t)$ tuples with ground truth evidence sentence indices $E_t = \{i_1, ..., i_m\}$. We consider two modifications to our model training, namely attention regularization \citep{Pruthi2020-1}, entropy maximization \citep{Feng2018}, and their combination. An illustration of both methods is shown in Figure \ref{fig:illustraction}.

\paragraph{Attention regularization} Attention regularization encourages our model $f(D, t)$ to leverage more information from $E_t$. Specifically, let $A = \{\alpha_1, ..., \alpha_n\}$ be the attention vector from the {\fontfamily{qcr}\selectfont [CLS]} token in the final layer to all tokens in $D$. During learning, we add the following loss to the training objective: $\ell_{attn} = - \log \sum_{i \in E_t} \alpha_{i}$, encouraging the model to attend to any token $i$ in the labeled sentence-level evidence set.

\paragraph{Entropy maximization} When there is no sufficient information contained in the text to infer any predictions, entropy maximization encourages a model to be uncertain, represented by a uniform probability distribution across all classes \citep{DeYoung2020, Feng2019}. Doing so should encourage the model to \emph{not} make predictions based on irrelevant sentences. We can achieve this by taking a reduced document $D' = D \setminus E_t$ as input by removing evidence $E_t$ from original document $D$. We treat $(D', t)$ pairs as extra training examples where we aim to maximize the entropy $- \sum_y P(y|D')\log P(y|D')$ over all possible $y$.\footnote{\liyan{We found this to work better than enforcing a uniform distribution over attention, which is much harder for the model to achieve.}}

\section{Experiments}

\subsection{Datasets and Evaluation Metrics}
\label{sec:data_eval}

We investigate our methods on (a) a small collection of brain MRI reports from radiologists’ observations; and (b) a modified version of the DocRED datatset. The statistics for both datatsets are included in Appendix \ref{section:statistics}. For both datasets, we evaluate on task accuracy (captured by either accuracy or prediction macro-F1) as well as evidence selection accuracy (macro-F1) or precision, measuring how well the model's evidence selection aligns with human annotations. We will use the \textsc{Sufficient} method defined in Section~\ref{sec:evi_extract} to select evidence sentences which guarantee that our predictions on the given evidence subsets will match the model's predictions on the full document. For the brain MRI report dataset, we evaluate evidence extraction by precision since human annotators typically only need to refer to one sentence to reach the conclusion but our model and baselines may extract more than one sentence.




\subsubsection{Brain MRI Reports} \label{sec:brainreport}

We present a new dataset of de-identified radiology reports from brain MRIs. It consists of the ``findings'' sections of reports, which present observations about the image, with labels for pre-selected key features by attending physicians and fellows. Crucially, these features are labeled \textbf{based on the original radiology image}, not the report. The document-level labels are therefore noisy because the radiologists' labels may disagree with the findings written in the report.

A key feature is an observable variable $t$, which can take on $d_t$ possible values. We focus on the evaluation of two key features, namely \emph{contrast enhancement} and \emph{mass effect}, since they appear in most of manually annotated reports. For our RoBERTa classification model, we only feed the document and train separate classifiers for each key feature, with no shared parameters between these. 



\paragraph{Annotation} We have a moderate number (327) of reports that have noisy labels from the process above. We treat these as our training set. However, all of these labels are document-level.

To evaluate models' performance on more fine-grained evidence labels, we randomly select $86$ unlabeled reports (not overlapping with the 327 for training) and asked four radiology residents to (1) assign key feature labels and reach consensus, while (2) highlighting sentences that support their decision making. We use Prodigy\footnote{\url{https://prodi.gy}} as our annotation interface. See Appendix~\ref{sec:instruction} for more details about our annotation instructions.

\paragraph{Pseudo sentence-level supervision} Since we have limited number of annotated reports for evaluation, we need a way to prepare weak sentence-level supervision ($E_t$) while training. To achieve this, we use sentences selected by our rule-based system as pseudo evidence to supervise models' behavior. We use 10\% of this as supervision while training for consistency with the DocRED setting.

\paragraph{Rule-based system} Our rule-based system uses keyword matching to identify instances of mass effect and contrast enhancement in the reports, and negspaCy 
to detect negations of these key features.

\paragraph{Data split} For the results in Section~\ref{Results}, we evaluate on reports that contain ground truth fine-grained annotations for either contrast enhancement or mass effect, respectively. There are 64 and 68 out of 86 documents total in each of these categories. We call this the \textsc{BrainMRI} set. When we restrict to this set for evaluation, all of the documents we study where the annotators labeled something related to \emph{contrast enhancement} end up having an explicit mention of it. However, for \emph{mass effect}, this is not always the case; Table~\ref{tab:reportexample} in Appendix shows an example where mass effect is discussed implicitly in the first sentence.

\subsubsection{Adapted DocRED}
\label{subsection: Docred}


DocRED \citep{Yao2019} is a document-level relation extraction (RE) dataset with large scale human annotation of relevant evidence sentences. Unlike sentence-level RE tasks \citep{Qin2018, alt2020}, it requires reading multiple sentences and reasoning about complex interactions between entities. We adapt this to a document-level relation classification task: a document $D$ and two entity mentions $e_1, e_2$ within the document are provided and the task is to predict the relation $r$ between $e_1$ and $e_2$. We synthesize these examples from the original dataset and sample random entity pairs from documents to which we assign an \emph{NA} class to construct negative pairs exhibiting no relation.

The model input is represented as: {\small\text{{\fontfamily{qcr}\selectfont[CLS]<ent-1>[SEP]<ent-2>[SEP]<doc>[SEP]}}}. \liyan{We use the encoding of \text{\fontfamily{qcr}\selectfont[CLS]} in the last layer to make predictions.}

To make the setting more realistic, we \emph{do not} use the large-scale evidence annotation and assume there is limited sentence-level supervision available. To be specific, we include 10\% fine-grained annotations in our adapted DocRED dataset.

\begin{table}
\small
\begin{tabular}{p{0.25\linewidth}  p{0.6\linewidth}}
\toprule
\textbf{Model Names} & \textbf{Input Text} \\ \midrule
\textsc{Direct} & None \\
\textsc{FullDoc} & Full document \\
\textsc{Ent} & Sentences containing at least one of the two query entities \\
\textsc{First2} & First two sentences \liyan{from a doc}. \\
\textsc{First3} & First three sentences \liyan{from a doc}.\\
\textsc{BestPair} & Two sentences yielding highest prediction prob. (incl.~variants using regularization) \\
\textsc{Sufficient}  & Sufficient sentences selected by interpretation methods  (incl.~variants using regularization) \\

\bottomrule
\end{tabular}
\caption{\label{notations} Model names used in the experiments and their associated evidence given as inputs.}
\end{table}


\subsection{Models}
\label{subsection:models}

Due to richer and higher-quality supervisions in the DocRED setting, we conduct a larger set of ablations and comparisons there. We compare against a subset of these models in the radiology setting.

\paragraph{Baselines} We consider a number of baselines for adapted DocRED which return both predicted labels and evidence. (1) \textsc{Direct} predicts the relation directly from the entity pairs without any sentences as input, using a model trained with just these inputs. (2) \textsc{FullDoc} takes the full document as selected evidence and uses the base RoBERTa model (3) \textsc{Ent} takes all sentences with entity mentions $e_1$ and $e_2$ as input; (4) \textsc{First2, First3} retrieve the first $2$ and $3$ sentences from a document, respectively; and (5) \textsc{BestPair} chooses the best sentence pair by first taking each individual sentence as input to the model and then picking top two sentences having highest probabilities on their predictions. 

\textsc{Sufficient} is our main method for both datasets, which we then augment with additional supervision as described in Section~\ref{sec:improving_ee}. We use subscripts {\fontfamily{qcr}\selectfont attn, entropy, both} and {\fontfamily{qcr}\selectfont none} to represent attention regularization, entropy maximization, the combination of two, and neither. \liyan{Both \textsc{BestPair} and \textsc{Sufficient} methods leverage backbone RoBERTa models trained with loss functions mentioned above, differing only in their evidence selection.}

Table \ref{notations} summarizes the abbreviated names of models and their inputs. Training details are described in Appendix \ref{section:trainingdetails}. 

\paragraph{Metrics} We report both the accuracy and F$_1$ for the model (\textbf{Full Doc}) as well as evaluation of \textbf{Evidence} selection compared to human judgments, either precision or F$_1$. We also report results in the \textbf{Reduced Doc} setting, where only the selected evidence sentences are fed to the RoBERTa model (trained over whole documents) as input. For our \textsc{Sufficient} method, this accuracy is the same as the full method by construction, but note that it can differ for other methods. This reduced setting serves as a sanity check for the faithfulness of our explanation techniques.

\liyan{Note once again that accuracy in the Full Doc case can differ for our methods that are trained with different regularization schemes, as these yield different models that return different predicted labels in addition to different evidence.}



\begin{table}
    \small
    \centering
    \renewcommand{\tabcolsep}{1.1mm}
  \begin{tabular}{lcccccc}
    \toprule
    \multirow{3}{*}{\textbf{Model}} &
     \multicolumn{4}{c}{\textbf{Label}} &
     \multicolumn{2}{c}{\textbf{Evidence}} \\ 
     &\multicolumn{2}{c}{\textbf{Full Doc}} &
      \multicolumn{2}{c}{\textbf{Reduced Doc}}& 
      \multicolumn{2}{c}{} \\ 
      & {\textbf{Acc}} & {\textbf{F1}} & {\textbf{Acc}} & {\textbf{F1}} & {\textbf{Pre}} & \textbf{Len}\\
      \midrule
      \multicolumn{7}{c}{Mass Effect} \\ \midrule
      \liyan{\textsc{FullDoc}} & 66.6 & 42.1 & 66.6 & 42.1 & 16.5 & 10.1 \\
      \liyan{\textsc{First2}} & $-$ & $-$ & 82.4 & 45.2 & 21.3 & 2.00\\
      \liyan{\textsc{First3}} & $-$ & $-$ & 82.4 & 45.2 & 24.0 & 3.00\\
      \textsc{Rule} & 77.9 & 11.8 &  77.9 & 11.8 & \textbf{\underline{84.8}} & 1.46\\
      \midrule 
      \liyan{\textsc{BestPair}\textsubscript{none}} & 66.6 & 42.1 & 82.4 & 52.2 & 24.3 & 2.00 \\
      \liyan{\textsc{BestPair}\textsubscript{both}} & 76.7 & \textbf{60.0} & 79.4 & 44.3 & \underline{50.7} & 2.00 \\
      \midrule
      \textsc{Sufficient}\textsubscript{none} & 66.6 & 42.1 & \multicolumn{2}{c}{\multirow{4}{1.5cm}{\centering Identical to Full Doc}} & 16.5 & 2.84 \\
      \textsc{Sufficient}\textsubscript{attn} & 69.2 & 47.6 & \multicolumn{2}{c}{} & 65.6 & 2.31 \\
      \textsc{Sufficient}\textsubscript{entropy} & 45.3 & 0.0 & \multicolumn{2}{c}{} & 15.8 & 2.50 \\
      \textsc{Sufficient}\textsubscript{both} & 76.7 & \textbf{60.0} & \multicolumn{2}{c}{} & \underline{77.8} & 1.51 \\
      \midrule
      \multicolumn{7}{c}{Contrast Enhancement}\\ \midrule
      \liyan{\textsc{FullDoc}} & 69.5 & 60.9 & 69.5 & 60.9 & 13.5 & 10.1 \\
      \liyan{\textsc{First2}} & $-$ & $-$ & 67.2 & 55.3 & 14.1 & 2.00\\
      \liyan{\textsc{First3}} & $-$ & $-$ & 70.3 & 62.4 & 14.6 & 3.00\\
      \textsc{Rule} & 68.8 & 56.5 & 68.8 & 56.5 & \textbf{\underline{87.1}} &1.67\\
      \midrule
      \liyan{\textsc{BestPair}\textsubscript{none}} & 69.5 & 60.9 & 73.4 & 67.7 & 10.9 & 2.00 \\
      \liyan{\textsc{BestPair}\textsubscript{both}} & \underline{90.8} & \textbf{\underline{87.2}} & \underline{89.1} & \underline{88.4} & \underline{54.7} & 2.00 \\
      \midrule
      \textsc{Sufficient}\textsubscript{none} & 69.5 & 60.9 & \multicolumn{2}{c}{\multirow{4}{1.5cm}{\centering Identical to Full Doc}} &  33.5 & 2.84 \\
      \textsc{Sufficient}\textsubscript{attn} & \underline{85.8} & \underline{81.0} & \multicolumn{2}{c}{} & \underline{60.7} & 2.48 \\
      \textsc{Sufficient}\textsubscript{entropy} & 71.5 & 59.5 & \multicolumn{2}{c}{} & 25.2 & 2.55 \\
      \textsc{Sufficient}\textsubscript{both} & \underline{90.8} & \textbf{\underline{87.2}} & \multicolumn{2}{c}{} & \underline{71.7} & 1.50 \\
    \bottomrule
  \end{tabular}
  \caption{\label{tab:brainmri} Model performance on \textsc{BrainMRI}. Models are evaluated under two settings by taking (a) full document (Full Doc); (b) selected evidence (Reduced Doc) as inputs.  \textsc{Rule} is the baseline mentioned in Section \ref{sec:brainreport}. \emph{Pre} stands for the precision of evidence selection, and \emph{Len} is the average number of sentences extracted. Underlined results are better than \textsc{Sufficient}\textsubscript{none} on the corresponding metric according to a paired bootstrap test with $p=0.05$.}
\end{table}

\section{Results}

\label{Results}



\subsection{Results on Brain MRI}

Table~\ref{tab:brainmri} shows the performance of our models and baselines in terms of label prediction and evidence extraction. \liyan{For each result, we perform a paired bootstrap test comparing to \textsc{Sufficient}\textsubscript{none}. We underline results that are better at a significance level of $p=0.05$ on the corresponding metrics.} In the \emph{mass effect} setting, our \textsc{Sufficient}\textsubscript{both} model achieves the highest evidence extraction precision of the learning-based models, \liyan{exceeds \textsc{FullDoc}, \textsc{First2/3}, and \textsc{BestPair} on the metric by a large margin,} and nearly matches that of the rule-based system. It is difficult to be more reliable than a rule-based system, which will nearly always make correctly-sourced predictions. But this model is able to \textbf{combine that reliability with the higher F$_1$ of a learned model}. Note that due to the high base rates of certain findings, we focus on F$_1$ instead of accuracy. We see a similar pattern on \emph{contrast enhancement} as well, although the evidence precision is lower in that case.

\begin{table}
\small
\centering
\renewcommand{\tabcolsep}{1.2mm}
\begin{tabular}{lcccccc}
\toprule
\multirow{3}{*}{\textbf{Model}}
 & \multicolumn{4}{c}{\textbf{Label}}
 & \multicolumn{2}{c}{\textbf{Evidence}} \\
 & \multicolumn{2}{c}{\textbf{Full Doc}}
 & \multicolumn{2}{c}{\textbf{Reduced Doc}}
 & \multicolumn{2}{c}{\textbf{}} \\
 & \textbf{Acc} & \textbf{F1} & \textbf{Acc} & \textbf{F1} & \textbf{F1} & \textbf{Len}\\
\midrule
\textsc{Direct} & $-$ & $-$ & 66.4 & 45.3 & $-$ & $-$ \\
\textsc{FullDoc} & 83.0 & 66.0 & 83.0 & 66.0 & 34.9  & 8.03\\
\textsc{First2} & $-$ & $-$ & 75.3 & 58.1 & 47.9  & 2.00 \\
\textsc{First3} & $-$ & $-$ & 77.5 & 60.7 & 44.6  & 3.00 \\
\textsc{Ent} & $-$ & $-$ & 82.4 & 65.4 & 61.5  & 3.93 \\ \midrule
\textsc{BestPair}\textsubscript{none} & 83.0 & 66.0 & 73.9 & 55.3 & 39.2  & 2.00\\
\textsc{BestPair}\textsubscript{attn} & 83.2 & 65.0 & 73.4 & 53.5 & 43.9  & 2.00\\
\textsc{BestPair}\textsubscript{entropy} & 81.8 & 64.2 & 78.5 & 58.2 & 52.3 & 2.00\\
\textsc{BestPair}\textsubscript{both} & 82.7 & 66.5 & 81.6 & 65.3 & 66.2 & 2.00\\ \midrule
\textsc{Sufficient}\textsubscript{none} & 83.0 & 66.0 & \multicolumn{2}{c}{\multirow{4}{1.5cm}{\centering Identical to Full Doc}} & 67.2  & 1.42\\
\textsc{Sufficient}\textsubscript{attn} & 83.2 & 65.0 & \multicolumn{2}{c}{} & \underline{70.3}  & 1.45\\
\textsc{Sufficient}\textsubscript{entropy} & 81.8 & 64.2 & \multicolumn{2}{c}{} & \underline{69.9} & \textbf{1.65}\\
\textsc{Sufficient}\textsubscript{both} & 82.7 & 66.5 & \multicolumn{2}{c}{} & \underline{\textbf{73.1}} & \textbf{1.65}\\ 
\midrule
human & $-$ & $-$ & $-$ & $-$ & $-$ & 1.59 \\
\bottomrule
\end{tabular}
\caption{\label{evidencecompare} Model performance on adapted DocRED. Models are evaluated under two settings as in \textsc{BrainMRI}. Underlined results are better than \textsc{Sufficient}\textsubscript{none} on the corresponding metric according to a paired bootstrap test with $p=0.05$.}
\end{table}

These results show that learning-based systems make accurate predictions in this domain, and that their evidence extraction can be improved with better training, even in spite of the small size of the training set. In section~\ref{docred_result}, we focus on the adapted DocRED setting, which allows us to examine our model's performance in a higher-data regime.

\paragraph{Attribution scores are more peaked at the occurrence of key terms.} We conduct analysis on how the attribution scores from \textsc{Sufficient}\textsubscript{both} are peaked around the correct evidence compare to that from \textsc{Sufficient}\textsubscript{none} using our manually annotated set \textsc{BrainMRI}. We compute the mean of the instance-wise average and maximum of the normalized attribution mass falling into a few explicit tokens: \emph{enhancement} for contrast enhancement and \emph{effect} for mass effect, which are common explicit indicators in the context of specified key features. The results in Table \ref{tab:attribution} show attribution scores being peaked around the correct terms, highlighting that these models can be guided to not only make correct predictions but attend to the right information. 

\begin{table}[]
    \small
    \centering
    \begin{tabular}{lcccc}
    \toprule
    \multirow{2}{*}{\textbf{Model}}
     & \multicolumn{2}{c}{\textbf{Mass Effect}}
     & \multicolumn{2}{c}{\textbf{Ctr. Enhance.}} \\
     & \textbf{Mean} & \textbf{Max} & \textbf{Mean} & \textbf{Max} \\ \midrule
     \textsc{Sufficient}\textsubscript{none} & 7.3 & 7.4 & 28.6  & 29.8 \\
     \textsc{Sufficient}\textsubscript{both} & \textbf{18.9} & \textbf{19.2} & \textbf{37.9} & \textbf{42.0}  \\ 
    \bottomrule
    \end{tabular}
    \caption{Distributions of attribution mass over explicit cues (``enhancement'' for \emph{contrast enhancement} and ``effect'' for \emph{mass effect}) for our best model and the baseline. Mean/Max is the mean of instance-wise average/maximum of the normalized attribution mass falling on the given token.\vspace{-0.2in}}
    \label{tab:attribution}
\end{table}

Table \ref{tab:reportexample} in the Appendix shows visualizations of attribution scores for an example in \textsc{BrainMRI} using DeepLIFT. Even though baseline models make correct predictions, their attribution mass is diffused over the document. With the help of regularization, our model is capable of capturing implicit cues such as \emph{downward displacement of the brain stem}, although it is trained on an extremely small training set with only explicit cues like \emph{mass effect} in a weak sentence-level supervision framework. 


\subsection{Results on Adapted DocRED}\label{docred_result}

\paragraph{Comparison to baselines}
Table~\ref{evidencecompare} shows that the \textsc{Ent} baseline is quite strong at DocRED evidence extraction. However, our best method still exceeds this method on both label accuracy as well as evidence extraction while extracting more succinct explanations. We see that the ability to extract a variable-length explanation is key, with \textsc{First2}, \textsc{First3} and \textsc{BestPair} performing poorly. Notably, these methods exhibit a drop in accuracy in the reduced doc setting for each method compared to the full doc setting, showing that the explanations extracted are not faithful.

\paragraph{Learning-based models with appropriate regularization perform relatively better in this larger-data setting}  From Table \ref{tab:brainmri} and Table \ref{evidencecompare}, we can observe that various regularization techniques applied to \textsc{Sufficient} models maintain or improve overall model performance on both key feature and relation classification. We see that our \textsc{Sufficient} methods do not compromise on accuracy but make predictions based on plausible evidence sets, which is more evident when we have richer training data. We perform further error analysis in Appendix~\ref{sec:error_analysis}.

\paragraph{Faithfulness of techniques} One may be concerned that, like attention values \cite{Jain2019}, our feature attribution methods may not faithfully reflect the computation of the model. We emphasize again that the \textsc{Sufficient} paradigm on top of the DeepLIFT method \emph{is} faithful by our definition. For a model $f$, we measure the faithfulness by checking the agreement between $\hat{y} = f(D,t)$ and $y' = f(\hat{E}_t,t)$, where $\hat{E}_t$ is the extracted evidence we feed into the same model under the reduced document setting. This is shown for all methods in the ``Reduced doc'' columns in Tables~\ref{tab:brainmri} and \ref{evidencecompare}. We see a drop in performance from techniques such as \textsc{BestPair}: the full model does not make the same judgment on these evidence subsets, but by definition it does in the \textsc{Sufficient} setting.

As further evidence of faithfulness, we note that only a relatively small number of evidence sentences, in line with human annotations, are extracted in the \textsc{Sufficient} method.
These small subsets are indicated by feature attribution methods \emph{and} sufficient to reproduce the original model predictions with high confidence, suggesting that these explanations are faithful.

\section{Conclusion}

In this work, we develop techniques to employ small amount of \label{sentence-annotated} data to improve reliability of document-level IE systems in two domains. We systematically evaluate our model from perspectives of faithfulness and plausibility and show that we can substantially improve models' capability in focusing on supporting evidence while maintaining their predictive performance, leading to models that are ``right for the right reasons.''

\section*{Acknowledgments}

This work was partially supported by NSF Grant IIS-1814522 and a Texas Health Catalyst grant. Thanks to Scott Rudkin, Gregory Mittl, Raghav Mattay, and Chuan Liang for assistance with the annotation.

\bibliography{anthology,custom}

\begin{thebibliography}{53}
\expandafter\ifx\csname natexlab\endcsname\relax\def\natexlab#1{#1}\fi

\bibitem[{Alt et~al.(2020)Alt, Gabryszak, and Hennig}]{alt2020}
Christoph Alt, Aleksandra Gabryszak, and Leonhard Hennig. 2020.
\newblock \href {https://arxiv.org/abs/2004.08134} {Probing linguistic features
  of sentence-level representations in neural relation extraction}.
\newblock In \emph{Proceedings of ACL}.

\bibitem[{Bahdanau et~al.(2015)Bahdanau, Cho, and Bengio}]{Bahdanau2015}
Dzmitry Bahdanau, Kyunghyun Cho, and Yoshua Bengio. 2015.
\newblock Neural machine translation by jointly learning to align and
  translate.
\newblock In \emph{Proceedings of the International Conference on Learning
  Representations (ICLR)}.

\bibitem[{Cho et~al.(2017)Cho, Choi, and Lee}]{Cho2017}
Hyejin Cho, Wonjun Choi, and Hyunju Lee. 2017.
\newblock \href {https://doi.org/10.1186/s12859-017-1857-8} {A method for named
  entity normalization in biomedical articles: application to diseases and
  plants}.
\newblock \emph{{BMC} Bioinformatics}, 18(1).

\bibitem[{Christopoulou et~al.(2019)Christopoulou, Miwa, and
  Ananiadou}]{Christopoulou2019}
Fenia Christopoulou, Makoto Miwa, and Sophia Ananiadou. 2019.
\newblock \href {https://doi.org/10.18653/v1/d19-1498} {Connecting the dots:
  Document-level neural relation extraction with edge-oriented graphs}.
\newblock In \emph{Proceedings of the 2019 Conference on Empirical Methods in
  Natural Language Processing and the 9th International Joint Conference on
  Natural Language Processing ({EMNLP}-{IJCNLP})}. Association for
  Computational Linguistics.

\bibitem[{Cook et~al.(2018)Cook, Gee, Bryan, Duda, Chen, Botzolakis, Mohan,
  Rauschecker, Rudie, and Nasrallah}]{Cook2018}
Tessa Cook, James~C. Gee, R.~Nick Bryan, Jeffrey~T. Duda, Po-Hao Chen, Emmanuel
  Botzolakis, Suyash Mohan, Andreas Rauschecker, Jeffrey Rudie, and Ilya
  Nasrallah. 2018.
\newblock \href {https://doi.org/10.1117/12.2293691} {Bayesian network
  interface for assisting radiology interpretation and education}.
\newblock In \emph{Medical Imaging 2018: Imaging Informatics for Healthcare,
  Research, and Applications}. {SPIE}.

\bibitem[{Demner-Fushman et~al.(2009)Demner-Fushman, Chapman, and
  McDonald}]{DemnerFushman_cds}
Dina Demner-Fushman, Wendy~W. Chapman, and Clement~J. McDonald. 2009.
\newblock \href {https://doi.org/https://doi.org/10.1016/j.jbi.2009.08.007}
  {{What can natural language processing do for clinical decision support?}}
\newblock \emph{Journal of Biomedical Informatics}, 42(5):760--772.
\newblock Biomedical Natural Language Processing.

\bibitem[{Devlin et~al.(2019)Devlin, Chang, Lee, and Toutanova}]{Devlin2019}
Jacob Devlin, Ming-Wei Chang, Kenton Lee, and Kristina Toutanova. 2019.
\newblock \href {https://doi.org/10.18653/v1/N19-1423} {{BERT}: Pre-training of
  deep bidirectional transformers for language understanding}.
\newblock In \emph{Proceedings of the 2019 Conference of the North {A}merican
  Chapter of the Association for Computational Linguistics: Human Language
  Technologies, Volume 1 (Long and Short Papers)}, pages 4171--4186,
  Minneapolis, Minnesota. Association for Computational Linguistics.

\bibitem[{DeYoung et~al.(2020)DeYoung, Jain, Rajani, Lehman, Xiong, Socher, and
  Wallace}]{DeYoung2020}
Jay DeYoung, Sarthak Jain, Nazneen~Fatema Rajani, Eric Lehman, Caiming Xiong,
  Richard Socher, and Byron~C. Wallace. 2020.
\newblock \href {https://doi.org/10.18653/v1/2020.acl-main.408} {{ERASER}: A
  benchmark to evaluate rationalized {NLP} models}.
\newblock In \emph{Proceedings of the 58th Annual Meeting of the Association
  for Computational Linguistics}. Association for Computational Linguistics.

\bibitem[{Doshi-Velez and Kim(2017)}]{DoshiVelez2017}
Finale Doshi-Velez and Been Kim. 2017.
\newblock Towards a rigorous science of interpretable machine learning.
\newblock \emph{arXiv: Machine Learning}.

\bibitem[{Dua et~al.(2020)Dua, Singh, and Gardner}]{DuaEtAl2020}
Dheeru Dua, Sameer Singh, and Matt Gardner. 2020.
\newblock \href {https://doi.org/10.18653/v1/2020.acl-main.497} {Benefits of
  intermediate annotations in reading comprehension}.
\newblock In \emph{Proceedings of the 58th Annual Meeting of the Association
  for Computational Linguistics}, pages 5627--5634, Online. Association for
  Computational Linguistics.

\bibitem[{Duong et~al.(2019)Duong, Rudie, Wang, Xie, Mohan, Gee, and
  Rauschecker}]{Duong2019}
M.T. Duong, J.D. Rudie, J.~Wang, L.~Xie, S.~Mohan, J.C. Gee, and A.M.
  Rauschecker. 2019.
\newblock \href {https://doi.org/10.3174/ajnr.a6138} {Convolutional neural
  network for automated {FLAIR} lesion segmentation on clinical brain {MR}
  imaging}.
\newblock \emph{American Journal of Neuroradiology}, 40(8):1282--1290.

\bibitem[{Feng et~al.(2019)Feng, Wallace, and Boyd-Graber}]{Feng2019}
Shi Feng, Eric Wallace, and Jordan Boyd-Graber. 2019.
\newblock \href {https://doi.org/10.18653/v1/p19-1554} {Misleading failures of
  partial-input baselines}.
\newblock In \emph{Proceedings of the 57th Annual Meeting of the Association
  for Computational Linguistics}. Association for Computational Linguistics.

\bibitem[{Feng et~al.(2018)Feng, Wallace, II, Iyyer, Rodriguez, and
  Boyd-Graber}]{Feng2018}
Shi Feng, Eric Wallace, Alvin~Grissom II, Mohit Iyyer, Pedro Rodriguez, and
  Jordan Boyd-Graber. 2018.
\newblock \href {https://doi.org/10.18653/v1/d18-1407} {Pathologies of neural
  models make interpretations difficult}.
\newblock In \emph{Proceedings of the 2018 Conference on Empirical Methods in
  Natural Language Processing}. Association for Computational Linguistics.

\bibitem[{Guoshun et~al.(2020)Guoshun, Zhijiang, Ivan, and Wei}]{nan2020lsr}
Nan Guoshun, Guo Zhijiang, Sekulić Ivan, and Lu~Wei. 2020.
\newblock Reasoning with latent structure refinement for document-level
  relation extraction.
\newblock In \emph{Proceedings of ACL}.

\bibitem[{Gururangan et~al.(2020)Gururangan, Marasović, Swayamdipta, Lo,
  Beltagy, Downey, and Smith}]{Gururangan2020}
Suchin Gururangan, Ana Marasović, Swabha Swayamdipta, Kyle Lo, Iz~Beltagy,
  Doug Downey, and Noah~A. Smith. 2020.
\newblock Don't stop pretraining: Adapt language models to domains and tasks.
\newblock In \emph{Proceedings of ACL}.

\bibitem[{Han et~al.(2020)Han, Gao, Lin, Peng, Yang, Xiao, Liu, Li, Zhou, and
  Sun}]{Han2020}
Xu~Han, Tianyu Gao, Yankai Lin, Hao Peng, Yaoliang Yang, Chaojun Xiao, Zhiyuan
  Liu, Peng Li, Jie Zhou, and Maosong Sun. 2020.
\newblock \href {https://aclanthology.org/2020.aacl-main.75} {More data, more
  relations, more context and more openness: A review and outlook for relation
  extraction}.
\newblock In \emph{Proceedings of the 1st Conference of the Asia-Pacific
  Chapter of the Association for Computational Linguistics and the 10th
  International Joint Conference on Natural Language Processing}, pages
  745--758, Suzhou, China. Association for Computational Linguistics.

\bibitem[{Hechtlinger(2016)}]{hechtlinger2016}
Yotam Hechtlinger. 2016.
\newblock \href {http://arxiv.org/abs/1611.07634} {Interpretation of prediction
  models using the input gradient}.
\newblock In \emph{Proceedings of the NeurIPS 2016 Workshop on Interpretable
  Machine Learning in Complex Systems}.

\bibitem[{Irvin et~al.(2019)Irvin, Rajpurkar, Ko, Yu, Ciurea-Ilcus, Chute,
  Marklund, Haghgoo, Ball, Shpanskaya, Seekins, Mong, Halabi, Sandberg, Jones,
  Larson, Langlotz, Patel, Lungren, and Ng}]{Irvin2019}
Jeremy Irvin, Pranav Rajpurkar, Michael Ko, Yifan Yu, Silviana Ciurea-Ilcus,
  Chris Chute, Henrik Marklund, Behzad Haghgoo, Robyn Ball, Katie Shpanskaya,
  Jayne Seekins, David Mong, Safwan Halabi, Jesse Sandberg, Ricky Jones, David
  Larson, Curtis Langlotz, Bhavik Patel, Matthew Lungren, and Andrew Ng. 2019.
\newblock \href {https://doi.org/10.1609/aaai.v33i01.3301590} {Chexpert: A
  large chest radiograph dataset with uncertainty labels and expert
  comparison}.
\newblock \emph{Proceedings of the AAAI Conference on Artificial Intelligence},
  33:590--597.

\bibitem[{Jacovi and Goldberg(2020)}]{Jacovi2020}
Alon Jacovi and Yoav Goldberg. 2020.
\newblock \href {https://doi.org/10.18653/v1/2020.acl-main.386} {Towards
  faithfully interpretable {NLP} systems: How should we define and evaluate
  faithfulness?}
\newblock In \emph{Proceedings of the 58th Annual Meeting of the Association
  for Computational Linguistics}. Association for Computational Linguistics.

\bibitem[{Jain and Wallace(2019)}]{Jain2019}
Sarthak Jain and Byron~C. Wallace. 2019.
\newblock \href {https://doi.org/10.18653/v1/N19-1357} {{A}ttention is not
  {E}xplanation}.
\newblock In \emph{Proceedings of the 2019 Conference of the North {A}merican
  Chapter of the Association for Computational Linguistics: Human Language
  Technologies, Volume 1 (Long and Short Papers)}, pages 3543--3556,
  Minneapolis, Minnesota. Association for Computational Linguistics.

\bibitem[{Jain et~al.(2020)Jain, Wiegreffe, Pinter, and Wallace}]{Jain2020}
Sarthak Jain, Sarah Wiegreffe, Yuval Pinter, and Byron~C. Wallace. 2020.
\newblock \href {https://doi.org/10.18653/v1/2020.acl-main.409} {Learning to
  faithfully rationalize by construction}.
\newblock In \emph{Proceedings of the 58th Annual Meeting of the Association
  for Computational Linguistics}. Association for Computational Linguistics.

\bibitem[{Lafferty et~al.(2001)Lafferty, McCallum, and Pereira}]{Lafferty2001}
John~D. Lafferty, Andrew McCallum, and Fernando C.~N. Pereira. 2001.
\newblock Conditional random fields: Probabilistic models for segmenting and
  labeling sequence data.
\newblock In \emph{Proceedings of the Eighteenth International Conference on
  Machine Learning}, ICML '01, page 282–289, San Francisco, CA, USA. Morgan
  Kaufmann Publishers Inc.

\bibitem[{Lei et~al.(2016)Lei, Barzilay, and Jaakkola}]{Lei2016}
Tao Lei, Regina Barzilay, and Tommi Jaakkola. 2016.
\newblock \href {https://doi.org/10.18653/v1/d16-1011} {Rationalizing neural
  predictions}.
\newblock In \emph{Proceedings of the 2016 Conference on Empirical Methods in
  Natural Language Processing}. Association for Computational Linguistics.

\bibitem[{Li et~al.(2016{\natexlab{a}})Li, Sun, Johnson, Sciaky, Wei, Leaman,
  Davis, Mattingly, Wiegers, and Lu}]{Li2016-2}
Jiao Li, Yueping Sun, Robin~J. Johnson, Daniela Sciaky, Chih-Hsuan Wei, Robert
  Leaman, Allan~Peter Davis, Carolyn~J. Mattingly, Thomas~C. Wiegers, and
  Zhiyong Lu. 2016{\natexlab{a}}.
\newblock \href {https://doi.org/10.1093/database/baw068} {{BioCreative} v
  {CDR} task corpus: a resource for chemical disease relation extraction}.
\newblock \emph{Database}, 2016:baw068.

\bibitem[{Li et~al.(2016{\natexlab{b}})Li, Monroe, and Jurafsky}]{Li2016-1}
Jiwei Li, Will Monroe, and Dan Jurafsky. 2016{\natexlab{b}}.
\newblock \href {http://arxiv.org/abs/1612.08220} {Understanding neural
  networks through representation erasure}.
\newblock \emph{ArXiv}, abs/1612.08220.

\bibitem[{Liu et~al.(2019)Liu, Ott, Goyal, Du, Joshi, Chen, Levy, Lewis,
  Zettlemoyer, and Stoyanov}]{liu2019}
Yinhan Liu, Myle Ott, Naman Goyal, Jingfei Du, Mandar Joshi, Danqi Chen, Omer
  Levy, Mike Lewis, Luke Zettlemoyer, and Veselin Stoyanov. 2019.
\newblock {RoBERTa: A Robustly Optimized BERT Pretraining Approach}.
\newblock In \emph{arXiv}.

\bibitem[{Loshchilov and Hutter(2019)}]{Loshchilov2019}
Ilya Loshchilov and Frank Hutter. 2019.
\newblock \href {https://openreview.net/forum?id=Bkg6RiCqY7} {Decoupled weight
  decay regularization}.
\newblock In \emph{International Conference on Learning Representations}.

\bibitem[{McDermott et~al.(2020)McDermott, Hsu, Weng, Ghassemi, and
  Szolovits}]{Matthew2020}
Matthew B.~A. McDermott, Tzu{-}Ming~Harry Hsu, Wei{-}Hung Weng, Marzyeh
  Ghassemi, and Peter Szolovits. 2020.
\newblock \href {http://arxiv.org/abs/2006.15229} {Chexpert++: Approximating
  the chexpert labeler for speed, differentiability, and probabilistic output}.
\newblock \emph{CoRR}, abs/2006.15229.

\bibitem[{Miller(2019)}]{Miller2019}
Tim Miller. 2019.
\newblock \href {https://doi.org/10.1016/j.artint.2018.07.007} {Explanation in
  artificial intelligence: Insights from the social sciences}.
\newblock \emph{Artificial Intelligence}, 267:1--38.

\bibitem[{Mintz et~al.(2009)Mintz, Bills, Snow, and Jurafsky}]{mintz2009}
Mike Mintz, Steven Bills, Rion Snow, and Daniel Jurafsky. 2009.
\newblock \href {https://www.aclweb.org/anthology/P09-1113} {Distant
  supervision for relation extraction without labeled data}.
\newblock In \emph{Proceedings of the Joint Conference of the 47th Annual
  Meeting of the {ACL} and the 4th International Joint Conference on Natural
  Language Processing of the {AFNLP}}, pages 1003--1011, Suntec, Singapore.
  Association for Computational Linguistics.

\bibitem[{Paranjape et~al.(2020)Paranjape, Joshi, Thickstun, Hajishirzi, and
  Zettlemoyer}]{Paranjape2020}
Bhargavi Paranjape, Mandar Joshi, John Thickstun, Hannaneh Hajishirzi, and Luke
  Zettlemoyer. 2020.
\newblock \href {https://doi.org/10.18653/v1/2020.emnlp-main.153} {An
  information bottleneck approach for controlling conciseness in rationale
  extraction}.
\newblock In \emph{Proceedings of the 2020 Conference on Empirical Methods in
  Natural Language Processing ({EMNLP})}. Association for Computational
  Linguistics.

\bibitem[{Patel et~al.(2020)Patel, Konam, and Selvaraj}]{Patel2020}
Dhruvesh Patel, Sandeep Konam, and Sai~P. Selvaraj. 2020.
\newblock Weakly supervised medication regimen extraction from medical
  conversations.
\newblock In \emph{ClinicalNLP@EMNLP}.

\bibitem[{Peng et~al.(2017)Peng, Poon, Quirk, Toutanova, and tau
  Yih}]{Peng2017}
Nanyun Peng, Hoifung Poon, Chris Quirk, Kristina Toutanova, and Wen tau Yih.
  2017.
\newblock \href {https://doi.org/10.1162/tacl_a_00049} {Cross-sentence n-ary
  relation extraction with graph {LSTMs}}.
\newblock \emph{Transactions of the Association for Computational Linguistics},
  5:101--115.

\bibitem[{Pons et~al.(2016)Pons, Braun, Hunink, and
  Kors}]{Pons_2016_nlp_in_rad}
Ewoud Pons, Loes M.~M. Braun, M.~G.~Myriam Hunink, and Jan~A. Kors. 2016.
\newblock \href {https://doi.org/10.1148/radiol.16142770} {Natural language
  processing in radiology: A systematic review}.
\newblock \emph{Radiology}, 279(2):329--343.
\newblock PMID: 27089187.

\bibitem[{Pruthi et~al.(2020)Pruthi, Dhingra, Neubig, and
  Lipton}]{Pruthi2020-2}
Danish Pruthi, Bhuwan Dhingra, Graham Neubig, and Zachary~C. Lipton. 2020.
\newblock \href {https://doi.org/10.18653/v1/2020.findings-emnlp.353} {Weakly-
  and semi-supervised evidence extraction}.
\newblock In \emph{Findings of the Association for Computational Linguistics:
  {EMNLP} 2020}. Association for Computational Linguistics.

\bibitem[{Pruthi et~al.(2021)Pruthi, Dhingra, Soares, Collins, Lipton, Neubig,
  and Cohen}]{Pruthi2020-1}
Danish Pruthi, Bhuwan Dhingra, Livio~Baldini Soares, M.~Collins, Zachary~C.
  Lipton, Graham Neubig, and William~W. Cohen. 2021.
\newblock \href {https://arxiv.org/abs/2012.00893} {Evaluating explanations:
  How much do explanations from the teacher aid students?}
\newblock \emph{Transactions of the Association for Computational Linguistics}.

\bibitem[{Qin et~al.(2018)Qin, Xu, and Wang}]{Qin2018}
Pengda Qin, Weiran Xu, and William~Yang Wang. 2018.
\newblock \href {https://doi.org/10.18653/v1/p18-1199} {Robust distant
  supervision relation extraction via deep reinforcement learning}.
\newblock In \emph{Proceedings of the 56th Annual Meeting of the Association
  for Computational Linguistics (Volume 1: Long Papers)}. Association for
  Computational Linguistics.

\bibitem[{Rauschecker et~al.(2020)Rauschecker, Rudie, Xie, Wang, Duong,
  Botzolakis, Kovalovich, Egan, Cook, Bryan, Nasrallah, Mohan, and
  Gee}]{Rauschecker2020}
Andreas~M. Rauschecker, Jeffrey~D. Rudie, Long Xie, Jiancong Wang, Michael~Tran
  Duong, Emmanuel~J. Botzolakis, Asha~M. Kovalovich, John Egan, Tessa~C. Cook,
  R.~Nick Bryan, Ilya~M. Nasrallah, Suyash Mohan, and James~C. Gee. 2020.
\newblock \href {https://doi.org/10.1148/radiol.2020190283} {Artificial
  intelligence system approaching neuroradiologist-level differential diagnosis
  accuracy at brain {MRI}}.
\newblock \emph{Radiology}, 295(3):626--637.

\bibitem[{Ribeiro et~al.(2016)Ribeiro, Singh, and Guestrin}]{Ribeiro2016}
Marco~Tulio Ribeiro, Sameer Singh, and Carlos Guestrin. 2016.
\newblock \href {https://doi.org/10.1145/2939672.2939778} {"why should i trust
  you?": Explaining the predictions of any classifier}.
\newblock In \emph{Proceedings of the 22nd ACM SIGKDD International Conference
  on Knowledge Discovery and Data Mining}, KDD '16, page 1135–1144, New York,
  NY, USA. Association for Computing Machinery.

\bibitem[{Rudie et~al.(2019)Rudie, Xie, Wang, Duda, Choi, Mattay, Chen, Bryan,
  Botzolakis, Nasrallah, Cook, Mohan, Gee, and Rauschecker}]{RudieEtAl2019}
Jeffrey Rudie, Long Xie, Jiancong Wang, Jeffrey Duda, Joshua Choi, Raghav
  Mattay, Po-Hao Chen, R~Nick Bryan, Emmanuel Botzolakis, Ilya Nasrallah, Tessa
  Cook, Suyash Mohan, James Gee, and Andreas Rauschecker. 2019.
\newblock {Artificial Intelligence System for Automated Brain MR Diagnosis
  Performs at Level of Academic Neuroradiologists and Augments Resident
  Performance}.
\newblock In \emph{Proceedings of the Society for Imaging Informatics in
  Medicine (SIIM)}.

\bibitem[{Rudie et~al.(2021)Rudie, Duda, Duong, Chen, Xie, Kurtz, Ware, Choi,
  Mattay, Botzolakis, Gee, Bryan, Cook, Mohan, Nasrallah, and
  Rauschecker}]{Rudie2021}
Jeffrey~D. Rudie, Jeffrey Duda, Michael~Tran Duong, Po-Hao Chen, Long Xie,
  Robert Kurtz, Jeffrey~B. Ware, Joshua Choi, Raghav~R. Mattay, Emmanuel~J.
  Botzolakis, James~C. Gee, R.~Nick Bryan, Tessa~S. Cook, Suyash Mohan, Ilya~M.
  Nasrallah, and Andreas~M. Rauschecker. 2021.
\newblock \href {https://doi.org/10.1007/s10278-021-00470-1} {Brain {MRI} deep
  learning and bayesian inference system augments radiology resident
  performance}.
\newblock \emph{Journal of Digital Imaging}, 34(4):1049--1058.

\bibitem[{Selvaraju et~al.(2017)Selvaraju, Cogswell, Das, Vedantam, Parikh, and
  Batra}]{Selvaraju2017}
Ramprasaath~R. Selvaraju, Michael Cogswell, Abhishek Das, Ramakrishna Vedantam,
  Devi Parikh, and Dhruv Batra. 2017.
\newblock \href {https://doi.org/10.1109/iccv.2017.74} {Grad-{CAM}: Visual
  explanations from deep networks via gradient-based localization}.
\newblock In \emph{2017 {IEEE} International Conference on Computer Vision
  ({ICCV})}. {IEEE}.

\bibitem[{Shrikumar et~al.(2017)Shrikumar, Greenside, and
  Kundaje}]{Shrikumar2017}
Avanti Shrikumar, Peyton Greenside, and Anshul Kundaje. 2017.
\newblock Learning important features through propagating activation
  differences.
\newblock In \emph{Proceedings of the 34th International Conference on Machine
  Learning - Volume 70}, ICML'17, page 3145–3153. JMLR.org.

\bibitem[{Simonyan et~al.(2014)Simonyan, Vedaldi, and Zisserman}]{Simonyan2014}
Karen Simonyan, Andrea Vedaldi, and Andrew Zisserman. 2014.
\newblock \href {https://arxiv.org/pdf/1312.6034.pdf} {Deep inside
  convolutional networks: Visualising image classification models and saliency
  maps}.
\newblock In \emph{Workshop at International Conference on Learning
  Representations}.

\bibitem[{Smit et~al.(2020)Smit, Jain, Rajpurkar, Pareek, Ng, and
  Lungren}]{Smit2020}
Akshay Smit, Saahil Jain, Pranav Rajpurkar, Anuj Pareek, Andrew Ng, and Matthew
  Lungren. 2020.
\newblock \href {https://doi.org/10.18653/v1/2020.emnlp-main.117} {Combining
  automatic labelers and expert annotations for accurate radiology report
  labeling using {BERT}}.
\newblock In \emph{Proceedings of the 2020 Conference on Empirical Methods in
  Natural Language Processing (EMNLP)}, pages 1500--1519, Online. Association
  for Computational Linguistics.

\bibitem[{Strout et~al.(2019)Strout, Zhang, and
  Mooney}]{strout-etal-2019-human}
Julia Strout, Ye~Zhang, and Raymond Mooney. 2019.
\newblock \href {https://doi.org/10.18653/v1/W19-4807} {Do human rationales
  improve machine explanations?}
\newblock In \emph{Proceedings of the 2019 ACL Workshop BlackboxNLP: Analyzing
  and Interpreting Neural Networks for NLP}, pages 56--62, Florence, Italy.
  Association for Computational Linguistics.

\bibitem[{Sundararajan et~al.(2017)Sundararajan, Taly, and
  Yan}]{sundararajan2017}
Mukund Sundararajan, Ankur Taly, and Qiqi Yan. 2017.
\newblock \href {https://proceedings.mlr.press/v70/sundararajan17a.html}
  {Axiomatic attribution for deep networks}.
\newblock In \emph{Proceedings of the 34th International Conference on Machine
  Learning}, volume~70 of \emph{Proceedings of Machine Learning Research},
  pages 3319--3328. PMLR.

\bibitem[{Surdeanu et~al.(2012)Surdeanu, Tibshirani, Nallapati, and
  Manning}]{surdeanu2012}
Mihai Surdeanu, Julie Tibshirani, Ramesh Nallapati, and Christopher~D. Manning.
  2012.
\newblock \href {https://www.aclweb.org/anthology/D12-1042} {Multi-instance
  multi-label learning for relation extraction}.
\newblock In \emph{Proceedings of the 2012 Joint Conference on Empirical
  Methods in Natural Language Processing and Computational Natural Language
  Learning}, pages 455--465, Jeju Island, Korea. Association for Computational
  Linguistics.

\bibitem[{Wang et~al.(2018)Wang, Han, Lin, Liu, and Sun}]{wang2018}
Xiaozhi Wang, Xu~Han, Yankai Lin, Zhiyuan Liu, and Maosong Sun. 2018.
\newblock \href {https://www.aclweb.org/anthology/C18-1099} {Adversarial
  multi-lingual neural relation extraction}.
\newblock In \emph{Proceedings of the 27th International Conference on
  Computational Linguistics}, pages 1156--1166, Santa Fe, New Mexico, USA.
  Association for Computational Linguistics.

\bibitem[{Xiao et~al.(2020)Xiao, Yao, Xie, Han, Liu, Sun, Lin, and
  Lin}]{Xiao2020}
Chaojun Xiao, Yuan Yao, Ruobing Xie, Xu~Han, Zhiyuan Liu, Maosong Sun, Fen Lin,
  and Leyu Lin. 2020.
\newblock \href {https://doi.org/10.18653/v1/2020.emnlp-main.300} {Denoising
  relation extraction from document-level distant supervision}.
\newblock In \emph{Proceedings of the 2020 Conference on Empirical Methods in
  Natural Language Processing ({EMNLP})}. Association for Computational
  Linguistics.

\bibitem[{Yao et~al.(2019)Yao, Ye, Li, Han, Lin, Liu, Liu, Huang, Zhou, and
  Sun}]{Yao2019}
Yuan Yao, Deming Ye, Peng Li, Xu~Han, Yankai Lin, Zhenghao Liu, Zhiyuan Liu,
  Lixin Huang, Jie Zhou, and Maosong Sun. 2019.
\newblock \href {https://doi.org/10.18653/v1/p19-1074} {{DocRED}: A large-scale
  document-level relation extraction dataset}.
\newblock In \emph{Proceedings of the 57th Annual Meeting of the Association
  for Computational Linguistics}. Association for Computational Linguistics.

\bibitem[{Zhang et~al.(2016)Zhang, Marshall, and Wallace}]{Zhang2016}
Ye~Zhang, Iain Marshall, and Byron~C. Wallace. 2016.
\newblock \href {https://doi.org/10.18653/v1/d16-1076} {Rationale-augmented
  convolutional neural networks for text classification}.
\newblock In \emph{Proceedings of the 2016 Conference on Empirical Methods in
  Natural Language Processing}. Association for Computational Linguistics.

\bibitem[{Zhou et~al.(2021)Zhou, Huang, Ma, and Huang}]{zhou2021}
Wenxuan Zhou, Kevin Huang, Tengyu Ma, and Jing Huang. 2021.
\newblock Document-level relation extraction with adaptive thresholding and
  localized context pooling.
\newblock In \emph{Proceedings of the AAAI Conference on Artificial
  Intelligence}.

\end{thebibliography}
\bibliographystyle{acl_natbib}

\clearpage

\appendix


\section{Implementation Details}
\label{section:trainingdetails}

We train all RoBERTa models for 15 epochs with early stopping using $1$ TITAN-Xp GPU. We use AdamW \cite{Loshchilov2019} as our optimizer and initialize the model with {\fontfamily{qcr}\selectfont roberta-base} for DocRED and {\fontfamily{qcr}\selectfont biomed-roberta-base} \citep{Gururangan2020} for brain MRI data, both with $125$M parameters. The batch size is set to 16 for RoBERTa models trained with both attention regularization and entropy maximization and 8 for models with other loss functions, and the learning rate is 1e-5 with linear schedule warmup. 

The maximum number of tokens in each document is capped at 296 for modified DocRED and 360 for radiology reports. These numbers are chosen such that the number of tokens for around 95\% of the documents is within these limits. \liyan{Remaining tokens are clipped from the input.} 
The hidden state of the {\fontfamily{qcr}\selectfont [CLS]} token from the final layer is fed as input to a linear projection head to make predictions. The average training time for each model is around 4 GPU hours.

\section{Interpretation Methods Comparison} \label{sec:interpret_compare}

\liyan{We evaluate four interpretation methods on \textsc{Sufficient}\textsubscript{none} and \textsc{Sufficient}\textsubscript{both} using adapted DocRED. These methods are widely used in the literature, namely Integrated Gradients, LIME, DeepLIFT, and Input Gradient, as discussed in Section~\ref{sec:evi_extract}. We compare their evidence extraction capabilities by selecting a wide range of $\lambda$, which controls the number of sentences to be selected.}

\begin{figure}
    \small
    \centering
    \includegraphics[scale=0.31,trim=0 230mm 0mm 15mm]{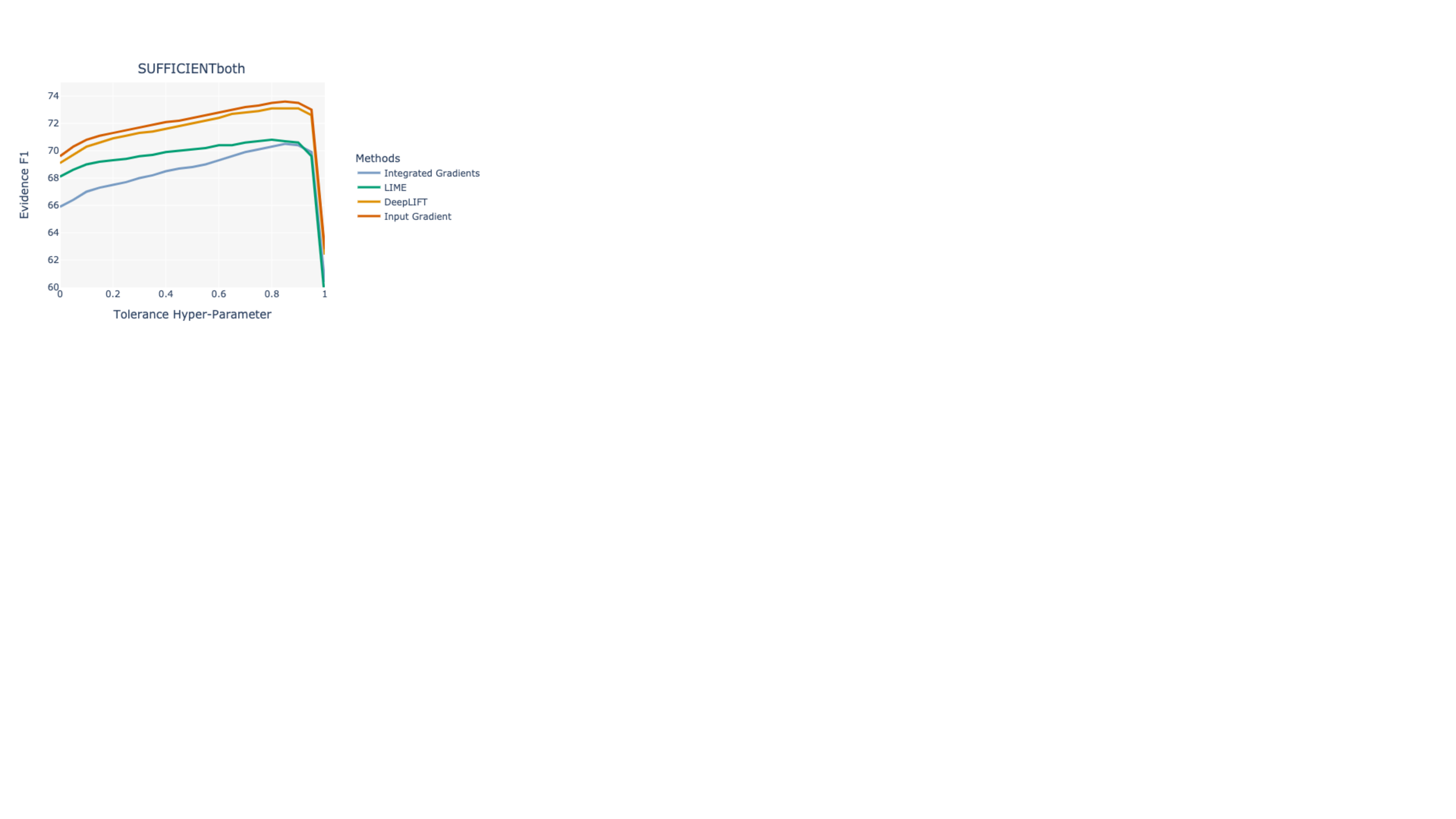}
    \caption{Evidence F1 on adapted DocRED under four model interpretation methods for \textsc{Sufficient}\textsubscript{both} over a wide range of $\lambda$.}
    \label{fig:hyper}
\end{figure}

\liyan{Results are shown in Figure~\ref{fig:hyper}. The four techniques generally perform similarly, with DeepLIFT and Input Gradient performing slightly better. For each interpretation method, the result of \textsc{Sufficient}\textsubscript{both} is significantly better than that of \textsc{Sufficient}\textsubscript{none}. Similar values of $\lambda$ between 0.8 and 0.9 (preferring to select more sentences) work well across all methods. Table~\ref{tab:lambda08} shows the comparison over the threshold ($\lambda=0.8$) we choose for our experiments in Section~\ref{Results}. In general, our method is robust to model interpretation techniques and evidence selection threshold $\lambda$.}

\liyan{Sentence ranking step mentioned in Section~\ref{sec:evi_extract} requires 0.3 GPU hour for Input Gradient and DeepLIFT, 2.5 GPU hours for Integrated Gradients, and 14 GPU hours for LIME. We choose 30 steps to approximate the integral for Integrated Gradients and 100 samples for each input to train the surrogate interpretable model (a linear model in our case) for LIME. }





\begin{table}[ht!]
\small
\begin{tabular}{lcc}
\toprule
 & \textsc{Sufficient}\textsubscript{none} & \textsc{Sufficient}\textsubscript{both} \\ \midrule
LIME & 54.0 & 70.8 \\
Integrated Gradients & 60.6 & 70.3 \\
DeepLIFT & 67.3 & 73.1 \\
Input Gradient & 68.3 & 73.5 \\
\bottomrule
\end{tabular}
\caption{Evidence F1 on adapted DocRED under four model interpretation methods for \textsc{Sufficient}\textsubscript{none} and \textsc{Sufficient}\textsubscript{both} when $\lambda = 0.8$.}
\label{tab:lambda08}
\end{table}

\section{Limitations and Risks}
\liyan{There are a few limitations of our work. First, we currently test our methods on document-level classification and slot-filling tasks, but there are other task formats like span extraction that we do not investigate here. Second, we focus on off-the-shelf pre-trained models (i.e. RoBERTa) in this paper, though we believe our methods could also be applied and adopted to other models. Finally, and most critically, the interpretation techniques we use are all fundamentally approximate; while visualizing model rationales can be useful in the context of clinical decision support systems, our evidence sets are not proof positive that a model's predictions are reliable. Such systems need to be carefully deployed to avoid misleading practitioners into trusting them too readily. We view this as the principal risk of our work.}

\section{Dataset statistics}
\label{section:statistics}

We provide the statistics for both adapted DocRED and brain MRI reports dataset in Table \ref{tab:statistics}. Both datasets are in English and the DocRED dataset is publicly available at \url{https://github.com/thunlp/DocRED}.

\section{Annotation Instructions} \label{sec:instruction}

\begin{figure*}
  \includegraphics[width=\textwidth]{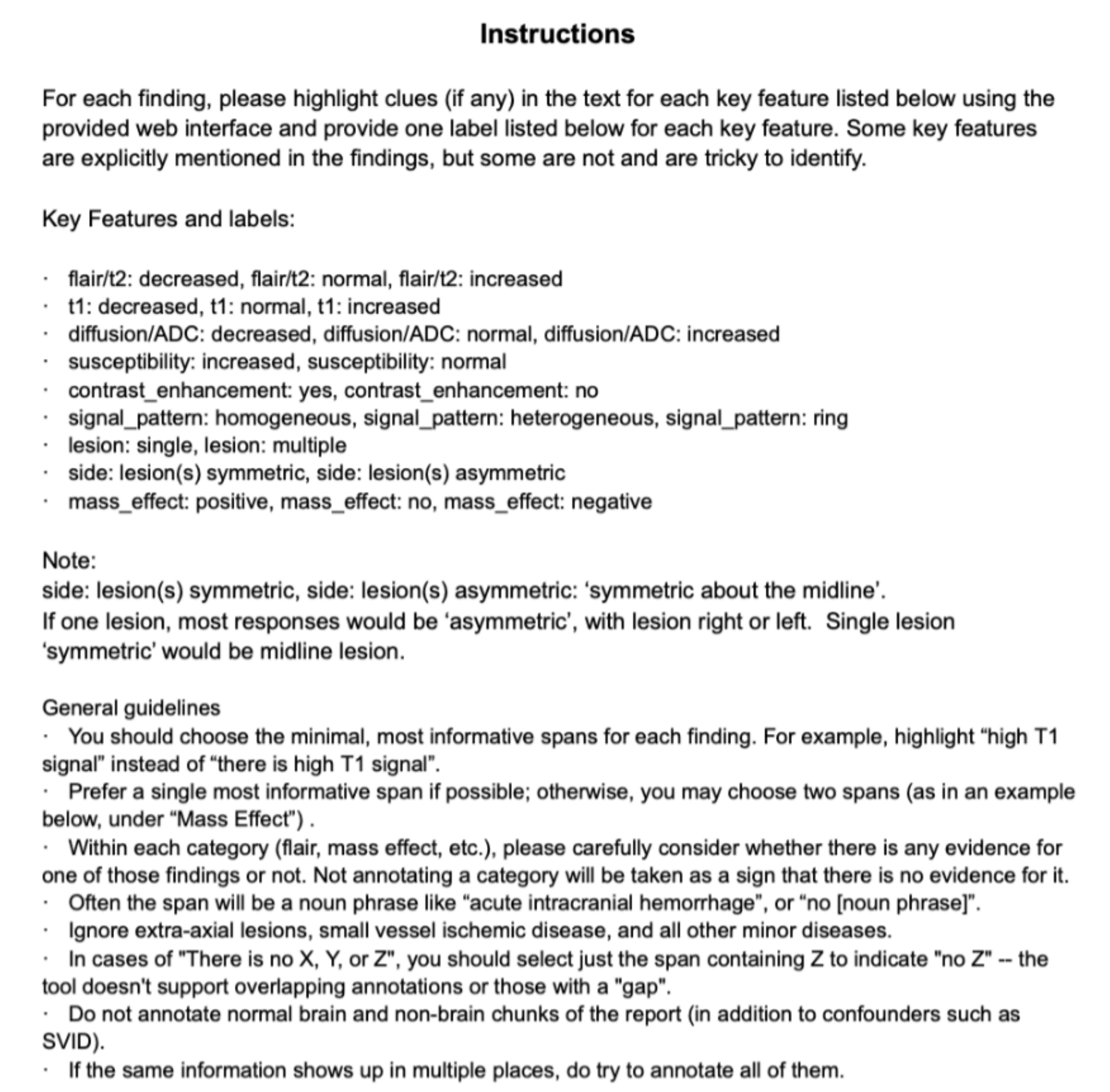}
  \caption{Annotation instructions.}
\label{fig:annotation_instructions}
\end{figure*}

We recruited four radiology residents to make annotations. They did not receive compensation for this project specifically. The annotation instructions for the BrainMRI dataset are provided in Figure~\ref{fig:annotation_instructions}. These were developed jointly with the annotators. In particular, decisions to exclude normal brain activity and confounders such as SVID were made to increase interannotator agreement after an initial round of annotation, making it easier for the labeling to focus on a single core disease or diagnosis per report.

\section{Error Analysis} \label{sec:error_analysis}

\begin{table*}[]\small
    \centering
    \begin{tabular}{lccccccc}
        \toprule
        \textbf{Dataset} & \textbf{Setting} & \textbf{\# doc.} & \textbf{\# inst.} & \textbf{\# word/inst.} & \textbf{\# sent./inst.} & \textbf{\# relation} & \textbf{\# NA\%}\\ \midrule
        \multirow{2}{*}{Adapted DocRED}
        & train & 3053 & 38180 & 203 & 8.1 & 96+1 & 33 \\
        & val & 1000 & 12323 & 203 & 8.1 & 96+1 & 33 \\ \midrule
        \multirow{2}{*}{Brain MRI}
        & train & 327 & 327 & 177 & 11.6 & $-$ & $-$\\
        & val & 86 & 86 & 132 & 10.1 & $-$ & $-$ \\ 
        \bottomrule
    \end{tabular}
    \caption{Statistics of the two document-level IE datasets. Each document may have multiple entity pairs of interest, giving rise to multiple instances in the adapted DocRED setting. For adapted DocRED, we have 96 relations from the data plus an \emph{NA} relation that we introduce for 1/3 of the data.}
    \label{tab:statistics}
\end{table*}

The first example in Table \ref{tab:cases} shows a representative case where our model predicts the correct relation and extracts reasonable supporting evidence. Unsurprisingly, this happens most often in simple cases when reasoning over the interaction of sentences is not required. 

We observe a few common types of errors. First, we see \textbf{potential alternatives for relations or evidence extraction}. From around $60\%$ of our randomly selected error cases, our model either predicts debatably correct relations or picks sentences that are related but not perfectly aligned with human annotations. The second row in Table \ref{tab:cases} illustrates an example where the two entities exhibit multiple relationships; the model's prediction is correct (\emph{Vienna} is place where \emph{Martinelli} was both born and died), but differs from the annotated ground truth and supporting evidence. Such relations are relatively frequent in this dataset; a more complex multi-label prediction format is necessary to fully support these.

Another type of error is \textbf{complex logical reasoning}. Even if our model can extract right evidence, it still fails in around $10\%$ of random error cases requiring sophisticated reasoning. For example, to correctly predict the relation between \emph{Theobald Tiger} and \emph{21 December 1935} in the third example in Table \ref{tab:cases}, a model needs to recognize that \emph{Theobald Tiger} and \emph{Kurt Tucholsky} are in fact the same entity by referring to \emph{pseudonym}, which is a challenging relation to recognize. 

Finally, the model sometimes \textbf{selects more sentences than we truly need}. Interestingly, this is an error in terms of \emph{evidence plausibility} but not in terms of \emph{prediction}. The number of extracted sentences is very high in around $25\%$ of the random error cases. The last row from Table \ref{tab:cases} is one of representative examples with this kind of error. Although our model possibly has already successfully extracted right evidence in the first two steps, it continues selecting unnecessary sentences because the prediction confidence is not high enough, a drawback in our way of selecting evidence mentioned in Section \ref{subsection:models}. Moreover, our model extracts one more sentence on average when predicting incorrect relations, suggesting that in these cases it does not cleanly focus on the correct information.

\begin{table*}[]
\centering
\small
\begin{tabular}{@{}cp{0.80\linewidth}@{}} \toprule
\textbf{Type} & \multicolumn{1}{c}{\textbf{Example}}\\ \midrule
  
    \multirow{3}{0.15\linewidth}{Predicts correctly and extracts right evidence}
    & \emph{[0]} Delphine ``Delphi'' Greenlaw is a fictional character on the \textbf{\textcolor{RedOrange}{New Zealand}} soap opera \textbf{\textcolor{Green}{Shortland Street}}, who was portrayed by Anna Hutchison between 2002 and 2004. ... \\
    \addlinespace[0.1cm]
    & \begin{tabular}{p{0.46\linewidth}p{0.46\linewidth}}
        \textbf{Predicted relation}: country of origin & \textbf{Relation}: country of origin\\
        \textbf{Extracted Evidence}: \emph{[0]} & \textbf{Annotated Evidence}:  \emph{[0]}\\
    \end{tabular} \\ \midrule

    \multirow{3}{0.15\linewidth}{Predicts debatably correct answer, extracts reasonable evidence}
    & \emph{[0]} \textbf{\textcolor{RedOrange}{Anton Erhard Martinelli}} (1684 – September 15 , 1747) was an Austrian architect and master - builder of Italian descent. \emph{[1]} \textbf{\textcolor{RedOrange}{Martinelli}} was born in \textbf{\textcolor{Green}{Vienna}}. ... \emph{[3]} \textbf{\textcolor{RedOrange}{Anton Erhard Martinelli}} supervised the construction of several important buildings in \textbf{\textcolor{Green}{Vienna}}, such as ... \emph{[4]} \textbf{\textcolor{RedOrange}{He}} designed ... \emph{[6]} \textbf{\textcolor{RedOrange}{He}} died in \textbf{\textcolor{Green}{Vienna}} in 1747. \\ 
    \addlinespace[0.1cm]
    & \begin{tabular}{p{0.46\linewidth}p{0.46\linewidth}}
        \textbf{Predicted relation}: place of birth & \textbf{Relation}: place of death\\
        \textbf{Extracted Evidence}: \emph{[1]} & \textbf{Annotated Evidence}:  \emph{[0, 6]}\\
    \end{tabular} \\ \midrule
    
    \multirow{3}{0.15\linewidth}{Predict incorrect example on examples requiring high amount of reasoning}
    & \emph{[0]} Kurt Tucholsky (9 January 1890 – \textbf{\textcolor{Green}{21 December 1935}}) was a German - Jewish journalist, satirist, and writer. \emph{[1]} He also wrote under the pseudonyms Kaspar Hauser (after the historical figure), Peter Panter, \textbf{\textcolor{RedOrange}{Theobald Tiger}} and Ignaz Wrobel. ... \\ 
    \addlinespace[0.1cm]
    & \begin{tabular}{p{0.46\linewidth}p{0.46\linewidth}}
        \textbf{Predicted relation}: \emph{NA} & \textbf{Relation}: date of death\\
        \textbf{Extracted Evidence}: \emph{[0]} & \textbf{Annotated Evidence}: \emph{[0]}\\
    \end{tabular} \\ \midrule
    
    \multirow{3}{0.15\linewidth}{Selecting more sentences than are needed}
    & \emph{[0]} \textbf{\textcolor{RedOrange}{Henri de Boulainvilliers}} ... was a \textbf{\textcolor{Green}{French}} nobleman, writer and historian. ... \emph{[2]} Primarily remembered as an early modern historian of the \textbf{\textcolor{Green}{French State}}, \textbf{\textcolor{RedOrange}{Boulainvilliers}} also published an early \textbf{\textcolor{Green}{French}} translation of Spinoza's Ethics and ... \emph{[3]} The \textbf{\textcolor{RedOrange}{Comte de Boulainvilliers}} traced his lineage to ... \emph{[5]} Much of \textbf{\textcolor{RedOrange}{Boulainvilliers}}' historical work ... \\
    \addlinespace[0.1cm]
    & \begin{tabular}{p{0.46\linewidth}p{0.46\linewidth}}
        \textbf{Predicted relation}: country of citizenship & \textbf{Relation}: country of citizenship\\
        \textbf{Extracted Evidence}: \emph{[2, 0, 1, 5, 4, 3]} & \textbf{Annotated Evidence}: \emph{[0, 2]}\\
    \end{tabular} \\

 \bottomrule 
\end{tabular}
\caption{Four types of representative examples that show models' behavior. In our adapted DocRED task, models are asked to predict relations among \textbf{\textcolor{RedOrange}{heads}} and \textbf{\textcolor{Green}{tails}}. Here we use model $\textsc{Sufficient}_\text{both}$ for illustrations, which has the best evidence extraction performance. Sentences in extracted evidence are ranked by DL.}
\label{tab:cases}
\end{table*}

\begin{table*}[]
\centering
\small
\begin{tabular}{@{}lp{0.85\linewidth}@{}} \toprule

\textbf{Model}  & \multicolumn{1}{c}{\textbf{An Example of \emph{mass effect}, \quad label: \emph{positive}, \quad evidence: 0 or 6}} \\ 
\midrule
\textsc{Sufficient}\textsubscript{none} & 
\textbf{[0] These images show \setlength{\fboxsep}{1.1pt}\colorbox{Red1}{evidence} of \setlength{\fboxsep}{1.1pt}\colorbox{Green1}{downward} displacement of the brain stem with \setlength{\fboxsep}{1.1pt}\colorbox{Green0}{collapse} of the interpeduncular cistern and caudal displacement of the mammary bodies typical for intracran\setlength{\fboxsep}{1.1pt}\colorbox{Green1}{ial} \setlength{\fboxsep}{1.1pt}\colorbox{Green0}{hypertension}\setlength{\fboxsep}{1.1pt}\colorbox{Red1}{.}} [1] There is diffuse pachymeningeal \setlength{\fboxsep}{1.1pt}\colorbox{Red0}{enhancement} evident\setlength{\fboxsep}{1.1pt}\colorbox{Red1}{.} [2] B\setlength{\fboxsep}{1.1pt}\colorbox{Green1}{ilateral} extra axial collections are evident the \setlength{\fboxsep}{1.1pt}\colorbox{Green0}{do not} conform to the \setlength{\fboxsep}{1.1pt}\colorbox{Red0}{imaging} \setlength{\fboxsep}{1.1pt}\colorbox{Green1}{characteristics} \setlength{\fboxsep}{1.1pt}\colorbox{Red0}{of} \setlength{\fboxsep}{1.1pt}\colorbox{Green3}{CS}F are seen over\setlength{\fboxsep}{1.1pt}\colorbox{Red0}{lying} the hemispheres. [3] These likely reflect blood t\setlength{\fboxsep}{1.1pt}\colorbox{Red0}{inged} hy\setlength{\fboxsep}{1.1pt}\colorbox{Red1}{g}\setlength{\fboxsep}{1.1pt}\colorbox{Red0}{rom}as and there does appear to \setlength{\fboxsep}{1.1pt}\colorbox{Green0}{be} \setlength{\fboxsep}{1.1pt}\colorbox{Red0}{a} \setlength{\fboxsep}{1.1pt}\colorbox{Green0}{blood products} in \setlength{\fboxsep}{1.1pt}\colorbox{Red0}{the} deep \setlength{\fboxsep}{1.1pt}\colorbox{Green1}{tendon} \setlength{\fboxsep}{1.1pt}\colorbox{Green0}{portion} of the right \setlength{\fboxsep}{1.1pt}\colorbox{Red2}{sided} \setlength{\fboxsep}{1.1pt}\colorbox{Green1}{collection} \setlength{\fboxsep}{1.1pt}\colorbox{Red1}{on} the patient\setlength{\fboxsep}{1.1pt}\colorbox{Green1}{'s} \setlength{\fboxsep}{1.1pt}\colorbox{Green0}{left} \setlength{\fboxsep}{1.1pt}\colorbox{Red3}{see} \setlength{\fboxsep}{1.1pt}\colorbox{Green1}{image} \setlength{\fboxsep}{1.1pt}\colorbox{Red1}{14} \setlength{\fboxsep}{1.1pt}\colorbox{Green0}{series} \colorbox{Red0}{2}. [4]\setlength{\fboxsep}{1.1pt}\colorbox{Red0}{There} \setlength{\fboxsep}{1.1pt}\colorbox{Green0}{does} appear to be a discrete linear subd\setlength{\fboxsep}{1.1pt}\colorbox{Green0}{ural he}matoma \setlength{\fboxsep}{1.1pt}\colorbox{Red1}{along} the right tentorial leaf. [5] \colorbox{Red3}{Sub}\setlength{\fboxsep}{1.1pt}\colorbox{Green1}{d}ural \setlength{\fboxsep}{1.1pt}\colorbox{Green0}{collection} is noted on both sides of the falx as well\setlength{\fboxsep}{1.1pt}\colorbox{Red1}{.} \textbf{[6] \setlength{\fboxsep}{1.1pt}\colorbox{Green0}{There is} \setlength{\fboxsep}{1.1pt}\colorbox{Red3}{mass} \setlength{\fboxsep}{1.1pt}\setlength{\fboxsep}{1.1pt}\setlength{\fboxsep}{1.1pt}\colorbox{Red2}{effect} at the level of the \setlength{\fboxsep}{1.1pt}\colorbox{Green1}{tentorial} inc\setlength{\fboxsep}{1.1pt}\colorbox{Red0}{isure} due to transtentorial her\setlength{\fboxsep}{1.1pt}\colorbox{Red0}{n}iation with \setlength{\fboxsep}{1.1pt}\colorbox{Red1}{deform}ity of the mid\setlength{\fboxsep}{1.1pt}\colorbox{Green1}{brain}\setlength{\fboxsep}{1.1pt}\colorbox{Red1}{.}} [7] There is no evidence \setlength{\fboxsep}{1.1pt}\colorbox{Red0}{an acute} \setlength{\fboxsep}{1.1pt}\colorbox{Green4}{inf}ar\setlength{\fboxsep}{1.1pt}\colorbox{Red0}{ct}\setlength{\fboxsep}{1.1pt}\colorbox{Red1}{.} [8] No parenchymal \setlength{\fboxsep}{1.1pt}\colorbox{Red0}{hemorrh}age is \setlength{\fboxsep}{1.1pt}\colorbox{Red0}{evident}\setlength{\fboxsep}{1.1pt}\colorbox{Green2}{.} [9] \setlength{\fboxsep}{1.1pt}\colorbox{Green3}{Apart} from the meningeal \setlength{\fboxsep}{1.1pt}\colorbox{Red0}{enhancement} there is no abnormal \setlength{\fboxsep}{1.1pt}\colorbox{Red0}{enhancement} noted.
\\ 
\midrule
\textsc{Sufficient}\textsubscript{both} & 
\textbf{[0] These images show \setlength{\fboxsep}{1.1pt}\colorbox{Red0}{evidence} of \setlength{\fboxsep}{1.1pt}\colorbox{Green4}{downward} \setlength{\fboxsep}{1.1pt}\colorbox{Green2}{displacement} of \setlength{\fboxsep}{1.1pt}\colorbox{Green2}{the} \setlength{\fboxsep}{1.1pt}\colorbox{Green3}{brain stem} with \setlength{\fboxsep}{1.1pt}\colorbox{Green3}{collapse} of the interpeduncular \setlength{\fboxsep}{1.1pt}\colorbox{Green0}{cis}tern and caudal displacement of the \setlength{\fboxsep}{1.1pt}\colorbox{Green0}{mamm}ary \setlength{\fboxsep}{1.1pt}\colorbox{Green0}{bodies} \setlength{\fboxsep}{1.1pt}\colorbox{Red1}{typical} for \setlength{\fboxsep}{1.1pt}\colorbox{Red1}{intr}ac\setlength{\fboxsep}{1.1pt}\colorbox{Green2}{ranial} \setlength{\fboxsep}{1.1pt}\colorbox{Green3}{hypertension}}. [1] There is diffuse pachymeningeal enhancement evident. [2] Bilateral extra axial collections are evident the do not conform to the imaging characteristics of \setlength{\fboxsep}{1.1pt}\colorbox{Green0}{CS}F are seen overlying the hemispheres. [3] These likely reflect blood tinged hyg\setlength{\fboxsep}{1.1pt}\colorbox{Red0}{rom}as and there does appear to be \setlength{\fboxsep}{1.1pt}\colorbox{Red0}{a} \setlength{\fboxsep}{1.1pt}\colorbox{Green0}{blood products} in the deep \setlength{\fboxsep}{1.1pt}\colorbox{Green0}{tendon} portion of the right sided \setlength{\fboxsep}{1.1pt}\colorbox{Green0}{collection} on the patient's left \setlength{\fboxsep}{1.1pt}\colorbox{Red0}{see} image 14 series 2. [4] There does appear to be a discrete linear subdural hematoma along the right tentorial leaf. [5] Subdural collection is noted on both sides of the falx as well. \textbf{[6] There is \setlength{\fboxsep}{1.1pt}\colorbox{Green4}{mass} \setlength{\fboxsep}{1.1pt}\colorbox{Red5}{effect} \setlength{\fboxsep}{1.1pt}\colorbox{Green2}{at} the \setlength{\fboxsep}{1.1pt}\colorbox{Red0}{level of} the tentorial inc\setlength{\fboxsep}{1.1pt}\colorbox{Green3}{isure} due to transtentorial her\setlength{\fboxsep}{1.1pt}\colorbox{Red2}{n}\setlength{\fboxsep}{1.1pt}\colorbox{Green3}{iation} with deformity of the \setlength{\fboxsep}{1.1pt}\colorbox{Green2}{midbrain}\setlength{\fboxsep}{1.1pt}\colorbox{Red1}{.}} [7] There is no \setlength{\fboxsep}{1.1pt}\colorbox{Red1}{evidence} an acute \setlength{\fboxsep}{1.1pt}\colorbox{Green4}{inf}arct. [8] No parenchymal hemorrhage is evident. [9] Apart from the meningeal enhancement there is no abnormal enhancement noted.
\\

 \bottomrule 
\end{tabular}
\caption{An illustration of models' attribution scores over a report from \textsc{BrainMRI} using DeepLift with and w/o regularization techniques. \textsc{Sufficient}\textsubscript{both} appears to leverage more information from right sentences.}
\label{tab:reportexample}
\end{table*}

\end{document}